\documentclass[10pt,twocolumn,twoside]{IEEEtran}
\usepackage{subfigure}
\usepackage{fixltx2e}
\usepackage{dblfloatfix}
\usepackage{amsmath}

\usepackage{array}
\usepackage{mdwmath}
\usepackage{mdwtab}
\usepackage{fixltx2e}
\usepackage{url}
\usepackage{color}
\usepackage[]{graphicx}
\usepackage{epsfig}
\usepackage{bbm}
\usepackage{enumerate}
\usepackage{amsfonts}
\usepackage{amsmath}
\usepackage{amssymb}
\usepackage{cite}
\usepackage{algorithm}
\usepackage{algpseudocode}
\usepackage{eqparbox}
\usepackage{comment}
\usepackage{epstopdf}

\newcommand{\vsaftfig}{\vspace{-0.15in}}
\newcommand{\vsbefsec}{\vspace{-0.05in}}

\makeatletter
\define@key{Gin}{ytimefigdim}[true]{%
    \edef\@tempa{{Gin}{width=0.55\columnwidth ,height=0.35\columnwidth ,clip=true ,trim=1in 0 0 1in}}%
    \expandafter\setkeys\@tempa
}
\define@key{Gin}{phcapdim}[true]{%
    \edef\@tempa{{Gin}{width=0.25\columnwidth, height=0.35\columnwidth, clip=true ,trim=0 0 0 0}}%
    \expandafter\setkeys\@tempa
}
\define@key{Gin}{capdim}[true]{%
    \edef\@tempa{{Gin}{width=0.25\columnwidth, height=0.35\columnwidth, clip=true ,trim=0 0 0 0}}%
    \expandafter\setkeys\@tempa
}
\define@key{Gin}{xyfigdim}[true]{%
    \edef\@tempa{{Gin}{width=0.35\columnwidth, height=0.35\columnwidth, clip=true ,trim=0 0 0 0 0}}%
    \expandafter\setkeys\@tempa
}

\define@key{Gin}{realxyfigdim}[true]{%
    \edef\@tempa{{Gin}{width=0.45\columnwidth, height=0.47\columnwidth ,clip=true ,trim={5.1in 4.5in 4.4in 4.5in}}}%
    \expandafter\setkeys\@tempa
}

\define@key{Gin}{realslicefigdim}[true]{%
    \edef\@tempa{{Gin}{width=0.68\columnwidth, height=0.45\columnwidth ,clip=true ,trim={0.3in 0.1in 0 0.5in}}}%
    \expandafter\setkeys\@tempa
}

\define@key{Gin}{sclxyfigdim}[true]{%
    \edef\@tempa{{Gin}{width=0.35\columnwidth, height=0.37\columnwidth ,clip=true ,trim={3.7in 3in 3in 3.0in}}}%
    \expandafter\setkeys\@tempa
}

\define@key{Gin}{plotdim}[true]{%
    \edef\@tempa{{Gin}{width=0.63\columnwidth, clip=true ,trim={0.05in 0 0.4in 0}}}%
    \expandafter\setkeys\@tempa
}

\makeatother

\include{defs}

\newcommand{\ETD}{\text{ETD}}

\IEEEoverridecommandlockouts

\begin{document}

\title{A Framework for Dynamic Image Sampling Based on Supervised Learning (SLADS)}

\author{\IEEEauthorblockN{G.~M.~Dilshan~P.~Godaliyadda\IEEEauthorrefmark{1},
Dong~Hye~Ye\IEEEauthorrefmark{1}, 
Michael~D.~Uchic\IEEEauthorrefmark{3}, 
Michael~A.~Groeber\IEEEauthorrefmark{3},
Gregery~T.~Buzzard\IEEEauthorrefmark{2},
and Charles~A.~Bouman\IEEEauthorrefmark{1}\\
}
\IEEEauthorblockA{\IEEEauthorrefmark{1}School of Electrical and Computer Engineering, 
Purdue University, West Lafayette, IN 47907, USA}\\
\IEEEauthorblockA{\IEEEauthorrefmark{2}Department of Mathematics, 
Purdue University, West Lafayette, IN 47907, USA}\\
\IEEEauthorblockA{\IEEEauthorrefmark{3}Air Force Research Laboratory, 
Materials and Manufacturing Directorate, Wright-Patterson AFB, OH 45433, USA}\\
\vsaftfig
}

\begin{scriptsize}
\end{scriptsize}\maketitle

\begin{abstract}

Sparse sampling schemes have the potential to dramatically reduce image acquisition time while simultaneously reducing radiation damage to samples. However, for a sparse sampling scheme to be useful it is important that we are able to reconstruct the underlying object with sufficient clarity using the sparse measurements. In dynamic sampling, each new measurement location is selected based on information obtained from previous measurements. Therefore, dynamic sampling schemes have the potential to dramatically reduce the number of measurements needed for high fidelity reconstructions. However, most existing dynamic sampling methods for point-wise measurement acquisition tend to be computationally expensive and are therefore too slow for practical applications.

In this paper, we present a framework for dynamic sampling based on machine learning techniques, which we call a supervised learning approach for dynamic sampling (SLADS). In each step of SLADS, the objective is to find the pixel that maximizes the expected reduction in distortion (ERD) given previous measurements. SLADS is fast because we use a simple regression function to compute the ERD, and it is accurate because the regression function is trained using data sets that are representative of the specific application. In addition, we introduce a method to terminate dynamic sampling at a desired level of distortion, and we extended the SLADS methodology to sample groups of pixels at each step. Finally, we present results on computationally-generated synthetic data and experimentally-collected data to demonstrate a dramatic improvement over state-of-the-art static sampling methods.

\vsbefsec
\end{abstract}
\section{Introduction}
\label{sec:introduction}

In conventional point-wise image acquisition, an object is measured by acquiring every measurement in a rectilinear grid. However, in certain imaging techniques one high fidelity pixel measurement can take up to $0.05$ to $5$ seconds, which translates to between $3.6-364$ hours of acquisition time for an image with $512 \times 512$ resolution. Examples of such imaging techniques include X-ray diffraction spectroscopy \cite{Garth}, high resolution electron back scatter diffraction
(EBSD) microscopy\cite{EBSD}, and Raman spectroscopy, which are of great importance in material science and chemistry

Sparse sampling offers the potential to dramatically reduce the time required to acquire an image. In sparse sampling a subset of all available measurements are acquired, and the full resolution image is reconstructed from this set of sparse measurements. In addition to reducing image acquisition time, sparse sampling also has the potential to reduce the exposure of the object/person being imaged to potentially harmful radiation. This is critically important when imaging biological samples using X-rays, electrons, or even optical photons \cite{Smith2009,Egerton2004399}. Another advantage of sparse sampling is that it reduces the amount of measurement data that must be stored. 

However, for a sparse sampling method to be useful, it is critical that the reconstruction made from the sparse set of samples allows for accurate reconstruction of the underlying object.  Therefore, the selection of sampling locations becomes critically important. All methods that researchers have proposed for sparse sampling can broadly be sorted into two primary categories: static and dynamic. 

Static sampling refers to any method that collects measurements in a predetermined order. Random sampling strategies such as in \cite{Hyrum13}, low-discrepancy sampling \cite{LDSampling} and uniformly spaced sparse sampling methods \cite{Aditya} are examples of static sparse sampling schemes. Static sampling methods can also be based on a model of the object being sampled such as in \cite{Mueller2011,wang2010variable}. In these methods knowledge of the object geometry and sparsity are used to pre-determine the measurement locations.

Alternatively, dynamic sampling refers to any method that adaptively determines the next measurement location based on information obtained from previous measurements. Dynamic sampling has the potential to produce a high fidelity image with fewer measurements because of the information available from previous measurements. Intuitively, the previous measurements provide a great deal of information about the best location for future measurements. 

Over the years, a wide variety of dynamic sampling methods have been proposed for many different applications. We categorize these dynamic sampling methods into three primary categories --- dynamic compressive sensing methods where measurements are unconstrained projections, dynamic sampling methods developed for specific applications where measurements are not point-wise measurements, and dynamic sampling methods developed for single pixel measurements. 

In dynamic compressive sensing methods the objective at each step is to find the measurement that reduces the entropy the most. In these methods the entropy is computed using the previous measurements and a model for the underlying data. Examples of such methods include \cite{seeger2008compressed,carson2012communications,Carin08}. However, in these methods the measurements are unconstrained projections and therefore cannot readily be generalized for point-wise sampling. 

The next category of dynamic sampling (DS) methods in the literature are those developed for specific applications where the measurements are not point-wise measurements. One example is \cite{Vanlier2012}, where the authors modify the optimal experimental design \cite{atkinson2007optimum} framework to incorporate dynamic measurement selection in a biochemical network. Another example is presented by Seeger et al. in \cite{Seeger10} to select optimal K-space spiral and line measurements for magnetic resonance imaging (MRI). Then Batenburg et al. in \cite{joost2012dynamic} presents another example for binary computed tomography, where in each step the measurement that maximizes the information gain is selected.

There are also a few DS methods developed specifically for point-wise measurements. One example is presented in \cite{merryman2005adaptive} by Kova{\v{c}}evi{\'c} et al. In this algorithm, an object is initially measured using a sparse rectilinear grid. Then, if the intensity of a pixel is above a certain threshold, the vicinity of that pixel is also measured. However, the threshold here is empirically determined and therefore is not robust for general applications. Another point-wise dynamic sampling method was proposed in \cite{godaliyaddaMBDS}. Here, in each step the pixel which when measured reduces the posterior variance the most is selected for measurement. The posterior variance is computed using samples generated from the posterior distribution using the Metropolis-Hastings algorithm. However, Monte-Carlo methods such as the Metropolis-Hastings algorithm can be very slow for cases where the dimensions of the random vector are large. Another shortcoming of this method is that it does not account for the change of conditional variance in the full image due to a new measurement. 

In this paper, we present an algorithm for point-wise dynamic sampling based on supervised learning techniques that we first presented in \cite{Godaliyadda2}. We call this algorithm a supervised learning approach for dynamic sampling (SLADS). In each step of the SLADS algorithm we select the pixel that maximizes the expected reduction in distortion ($ERD$) given previous measurements. We estimate the $ERD$ from previous measurements using a simple regression function that we train offline. As a result, we can compute the $ERD$ very fast during dynamic sampling. However, for our algorithm to be accurate we need to train this regression function with reduction in distortion ($RD$) values  resulting from many different measurements. In certain cases, where the image sizes are large,  computing sufficiently many entries for the training database can be very time consuming. To solve this problem we introduce an efficient training scheme that allows us to extract many entries for the training database with just one reconstruction. We empirically validate this approximation for small images and then detail a method to find the parameters needed for this approximation when dealing with larger images.  Then we introduce a stopping condition for dynamic sampling, which allows us to stop when a desired distortion level is reached, as opposed to stopping after a predetermined number of samples are acquired. Finally we extend our algorithm to incorporate group-wise sampling so that multiple measurements can be selected in each step of the algorithm. 

In the results section of this paper, we first empirically validate our approximation to the $ERD$ by performing experiments on $64 \times 64$ sized computationally generated EBSD images. Then we compare SLADS with state-of-the-art static sampling methods by sampling both simulated EBSD and real SEM images. We observe that with SLADS we can compute a new sample location very quickly (in the range of $1$ - $100 \ ms$), and can achieve the same reconstruction distortion as static sampling methods with dramatically fewer samples. Finally, we evaluate the performance of group-wise SLADS by comparing it to SLADS and to static sampling methods.
\section{Dynamic Sampling Framework}

In sparse image sampling
we  measure only a subset 
of all available pixels. 
However, for a sparse sampling technique to be useful,
the measured pixels should yield an accurate reconstruction
of the underlying object.
Dynamic sampling (DS) is a particular form of sparse sampling
in which each new measurement
is informed by all the previous pixel measurements.

To formulate the dynamic sampling problem,
we first denote the unknown image
formed by imaging the entire underlying object 
as $X\in \mathbb{R}^N$.
We then denote the value of the pixel at location $r\in \Omega$ by $X_r$,
where $\Omega$ is the set of all locations in the image.

Now, we assume that $k$ pixels have already been measured
at a set of locations $\mathcal{S} = \{ s^{(1)}, \cdots , s^{(k)} \}$.
We then represent the measurements 
and the corresponding locations 
as a $k\times 2$ matrix
$$
Y^{(k)} = 
\left[ 
\begin{array}{c}
s^{(1)}, X_{s^{(1)}} \\
\vdots \\
s^{(k)}, X_{s^{(k)}} 
\end{array}
\right] \ .
$$
From these measurements, $Y^{(k)}$, we can compute an estimate of the unknown image $X$.
We denote this best current estimate of the image as $\hat{X}^{(k)}$.

Now we would like to determine the next pixel location $s^{(k+1)}$ to measure.
If we select a new pixel location $s$ and measure its value $X_s$,
then we can presumably reconstruct a better estimate of $X$. 
We denote this improved estimate as $\hat{X}^{(k;s)}$.

Of course, our goal is to minimize the distortion between $X$ and $\hat{X}^{(k;s)}$,
which we denote by the following function
\begin{equation}
D ( X , \hat{X}^{(k;s)} ) = \sum_{r\in \Omega} D ( X_r , \hat{X}_r^{(k;s)} ) \ ,
\label{eqn:defn difference}
\end{equation}
where $D ( X_r , \hat{X}_r^{(k; s)} )$ is some scalar measure of distortion between the two pixels $X_r$ and $\hat{X}_r^{(k; s)}$.
Here, the function $D(\cdot , \cdot )$ may depend on the specific application or type of image.
For example we can let $D(a , b )=\vert a-b \vert^l$ where $l \in \mathbb{Z^{+}}$.

In fact, minimizing this distortion is equivalent to maximizing the reduction in the distortion
that occurs when we make a measurement. 
To do this, we define $R_r^{(k;s)}$ as the 
reduction-in-distortion at pixel location $r$
resulting from a measurement at location $s$.
\begin{eqnarray}
R^{(k;s)} = D ( X , \hat{X}^{(k)} ) - D ( X , \hat{X}^{(k;s)} ) \ .
\label{eqn:RD}
\end{eqnarray}
Notice that typically $R^{(k;s)}$ will be a positive quantity since we expect
that the distortion should reduce when we collect additional information with new measurements.
While this is typically true, in specific situations $R^{(k;s)}$ can actually be negative
since a particular measurement might mislead one into increasing the distortion.
Perhaps more importantly, we cannot know the value of $R^{(k;s)}$ because
we do not know $X$. 

Therefore, our real goal will be to minimize the expected value of $R^{(k;s)}$ given our current measurements.
We define the expected reduction-in-distortion ($ERD$) as 
\begin{equation}
\bar{R}^{(k;s)}= \mathbb{E} \left[ R^{(k;s)} \vert Y^{(k)} \right] \ .
\label{eqn:ERD}
\end{equation}
The specific goal of our greedy dynamic sampling algorithm will be to
select the pixel $s$ according to the following rule. 
\begin{equation}
s^{(k+1)}= \arg \max_{s \in \Omega } \left\lbrace \bar{R}^{(k;s)} \right\rbrace
\label{eqn:ideal sampling strategy}
\end{equation}
Intuitively, equation~(\ref{eqn:ideal sampling strategy}) selects the next pixel to maximize 
the expected reduction-in-distortion given all the available information $Y^{(k)}$.

Once $s^{(k+1)}$ is determined, we then form a new measurement matrix given by
\begin{equation}
Y^{(k+1)} =
\left[ 
\begin{array}{c}
Y^{(k)} \\
s^{(k+1)}, X_{s^{(k+1)}} 
\end{array}
\right] \ .
\label{eqn:add new measurement}
\end{equation}
We repeat this process recursively until the stopping condition 
discussed in Section~\ref{sec:stopping condition} is achieved.

In summary, the greedy dynamic sampling algorithm is given by the following iteration.
\begin{figure}[H]
\centering
\fbox{
 \addtolength{\linewidth}{-5\fboxsep}%
 \addtolength{\linewidth}{-5\fboxrule}%
 \begin{minipage}{\linewidth}
\begin{algorithmic}
\State{$k \leftarrow 0$}
\Repeat 
\State \normalsize
\State{$s^{(k+1)}= \arg \max_{s \in \Omega } \left\lbrace  \bar{R}^{(k;s)} \right\rbrace$} 
\State \tiny
\State  \normalsize {$Y^{(k+1)} =
\left[ 
\begin{array}{c}
Y^{(k)} \\
s^{(k+1)}, X_{s^{(k+1)}} 
\end{array}
\right] 
$
}
\State \tiny
\State \normalsize {$k \leftarrow k+1$}
\State \tiny \normalsize
\Until \normalsize {Desired fidelity is achieved}
\end{algorithmic}
 \end{minipage}
}
\end{figure}

In Section~\ref{sec:stopping condition}, we will introduce a stopping condition 
that can be used to set a specific expected quality level for the reconstruction.

\section{Supervised Learning Approach for Dynamic Sampling (SLADS)}
\label{sec:SLADS}

The challenge in implementing this greedy dynamic sampling method
is accurately determining the $ERD$ function, $\bar{R}^{(k;s)}$.
Our approach to solve this problem is to use supervised 
learning techniques to determine this function from training data.

More specifically, the supervised learning approach for dynamic sampling (SLADS)
will use an off-line training approach 
to learn the relationship between the $ERD$
and the available measurements, $Y$,
so that we can efficiently predict the $ERD$.
More specifically, we would like to fit a regression function $f^{\theta} (\cdot )$, 
so that
\begin{equation}
\bar{R}^{(s)}= f^{\theta}_{s}  ( Y ) \ .
\label{eqn:SLFunction}
\end{equation}
Here $f^{\theta}_{s}   ( \cdot )$ denotes a non-linear regression function
determined through supervised learning,
and $\theta$ is the parameter vector that must be estimated in the learning process.
Notice above that we have dropped the superscript $k$
since we would like to estimate a function that will work for any number of previously measured pixels.

Now, to estimate $f^{\theta} (\cdot )$ 
we must construct a training database 
containing multiple corresponding pairs of $( R^{(s)} , Y )$.
Here, $R^{(s)}=D(X,\hat{X}) -D(X,\hat{X}^{(s)})$, 
where $\hat{X}$ is the best estimate of $X$ 
computed using the measurements $Y$,
and $\hat{X}^{(s)}$ is the best estimate of $X$ 
computed using the measurements $Y$ along with an additional measurement at location $s$. 

Notice that since $R^{(s)} $ is the reduction-in-distortion, 
it requires knowledge of the true image $X$.
Since this is an off-line training procedure, $X$ is available,
and the regression function, $f^{\theta}_{s} ( Y )$,
will compute the required conditional expectation $\bar{R}^{(s)}$.
However, in order to compute $R^{(s)}$ for a single value of $s$,
we must compute two full reconstructions, both $\hat{X}$ and $\hat{X}^{(s)}$.
Since reconstruction can be computationally expensive,
this means that creating a large database can be very computationally expensive.
We will address this problem 
and propose a solution to it in Section~\ref{sec:efficient training}.

For our implementation of SLADS, 
the regression function $f^{\theta}_{s} \left( Y \right)$ 
will be a function of a row vector containing features extracted from $Y$.
More specifically, at each location $s$, 
a $p$-dimensional feature vector $V_s$ will be computed using $Y$ 
and used as input to the regression function.
The specific choices of features used in $V_s$ are listed in Section~\ref{sec:descriptors};
however, other choices are possible.

From this feature vector, 
we then compute the $ERD$ using a linear predictor with the following form:
\begin{equation}
\bar{R}^{(s)} = f^{\theta}_{s} \left( Y \right) = V_s\, \theta.
\label{eqn:SLFunction with theta}
\end{equation}
We can estimate the parameter $\theta$ by solving the following least-squares regression
\begin{equation}
\hat{\theta} = \arg \min_{\theta \in \mathbb{R}^p}  \Vert \textbf{R} - \textbf{V} \theta \Vert^2 \ ,
\label{eqn:least squares to find theta}
\end{equation}
where $\textbf{R}$ is an $n$-dimensional column vector formed by
\begin{equation}
\textbf{R} = 
\left[ 
\begin{array}{c}
R^{(s_1)} \\
\vdots \\
R^{(s_n)} 
\end{array} 
\right] \ ,
\label{eqn:big R training DB}
\end{equation}
and $\textbf{V}$ is given by
\begin{equation}
\textbf{V} = 
\left[ 
\begin{array}{c}
V_{s_1} \\
\vdots \\
V_{s_n} 
\end{array} 
\right] \ .
\label{eqn:big V training DB}
\end{equation}
So together $(\textbf{R},\textbf{V})$ consist of $n$ training pairs, $\{ ( R_{s_i}, V_{s_i} ) \}_{i=1}^{n}$,
that are extracted from training data during an off-line training procedure.
The parameter $\theta$ is then given by
\begin{equation}
\hat{\theta} = \left( \textbf{V}^t \textbf{V} \right)^{-1} \textbf{V}^t \textbf{R} \ .
\label{eqn:solution to theta}
\end{equation}

Once $\hat{\theta}$ is estimated,
we can use it to find the $ERD$
for each unmeasured pixel during dynamic sampling.
Hence, we find the $k+1^{th}$ location to measure by solving
\begin{equation}
s^{(k+1)} = \arg \max_{s \in \Omega } \left( V^{(k)}_s \hat{\theta} \right)\ ,
\label{eqn:SLADSUpdate}
\end{equation}
where $V^{(k)}_s$ denotes the feature vector extracted from the measurements $Y^{(k)}$ 
at location $s$.
It is important to note that this computation can be done very fast.
The pseudo code for SLADS 
is shown in Figure~(\ref{fig:SLADS pseudocode}).

\begin{figure}[!h]
\centering
\fbox{
 \addtolength{\linewidth}{-5\fboxsep}%
 \addtolength{\linewidth}{-5\fboxrule}%
 \begin{minipage}{\linewidth}
\begin{algorithmic}
{
\Function{$Y^{(K)}$ $\gets$ \textsc{SLADS}}{$Y^{(k)},\hat{\theta},k$}
\State \tiny
\State \normalsize $\mathcal{S} \gets \left\lbrace s^{(1)},s^{(2)},\hdots s^{(k)} \right\rbrace $
\State \tiny
\normalsize  \While  {Stopping condition not met} 
\State \tiny
\normalsize \For{\ $\forall s \in \left\lbrace \Omega \setminus \mathcal{S} \right\rbrace$} 
\State \tiny
\normalsize  \State  Extract $V^{(k)}_s$
\State \tiny
\State \normalsize $\bar{R}^{(k;s)}\leftarrow V^{(k)}_s\hat{\theta}$
\State \tiny
\normalsize \EndFor
\State \tiny
\State \normalsize $s^{(k+1)} =\arg \underset{s \in  \left\lbrace \Omega \setminus \mathcal{S} \right\rbrace }{\operatorname{ max}} \left( \bar{R}^{(k;s)} \right)$ 
\State \vspace{3mm}
\State \normalsize
$Y^{(k+1)} = 
\left[ 
\begin{array}{c}
Y^{(k)} \\
s^{(k+1)}, X_{s^{(k+1)}} 
\end{array}
\right] \ $
\State \vspace{3mm}
\State \normalsize $\mathcal{S} \gets \left\lbrace \mathcal{S} \cup s^{(k+1)}\right\rbrace$
\State \scriptsize
\State \normalsize $k \leftarrow k+1$
\State \tiny
\normalsize  \EndWhile \normalsize
\State \normalsize \tiny
\State \normalsize $K\gets k$
\State \tiny
\normalsize  \EndFunction}
\end{algorithmic}
 \end{minipage}
}
\vspace{0.5mm}
\caption{SLADS algorithm in pseudo code. 
The inputs to the function are 
the initial measurements $Y^{(k)}$, 
the coefficients needed to compute the $ERD$, found in training, $\hat{\theta}$,
and $k$, the number of measurements.
When the stopping condition is met, 
the function will output the selected set of measurements $Y^{(K)}$.}
\label{fig:SLADS pseudocode}
\end{figure}

\subsection{Training for SLADS}
\label{sec:efficient training}

In order to train the SLADS algorithm,
we must form a large training database containing corresponding pairs of $R^{(s)}$ and $V_s$.
To do this, we start by selecting $M$ training images denoted by
$\left\lbrace X_1, X_2, \hdots , X_M \right\rbrace$. 
We also select a set of sampling densities represented by $p_1\%, p_2\%, \hdots , p_H\%$.

For image $X_m$ and each sampling density, $p_h\%$, 
we randomly select $p_h\%$ of all pixels in the image to represent the simulated measurement locations.
Then for each of the remaining unmeasured locations, $s$, in the image $X_m$,
we compute the pairs $( R^{(s)},V_s )$ and save them to the training database. 
This process is then repeated for all the sampling densities and all the images
to form a complete training database.

\begin{figure*}
\centering
\includegraphics[scale=0.5]{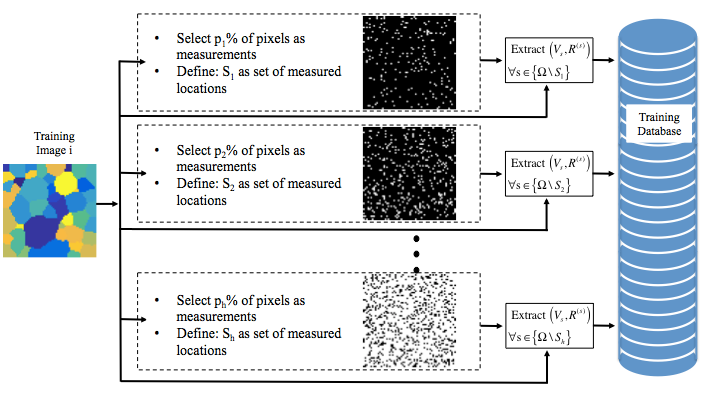}
\caption{Illustration of how training data is extracted  
from one image in the training database. 
We first select $p_1\%$ of the pixels in the image 
and consider them as measurements $Y$. 
Then for all unmeasured pixel locations ($s \in \left\lbrace \Omega \setminus \mathcal{S}_1 \right\rbrace $)
we extract a feature vector $V_s$ 
and also compute $R^{(s)}$. 
We the repeat the process for when $p_2, p_3, \hdots $ and $p_h\%$
of the pixels are considered measurements.
Here again $\Omega$ is the set of all locations in the training image
and $\mathcal{S}_i$ is the set of measured locations
when $p_i \%$ of pixels are selected as measurements. 
All these pairs of $\left( V_s,R^{(s)} \right)$ 
are then stored in the training database.}
\label{fig:building training DB}
\end{figure*}

Figure~\ref{fig:building training DB} illustrates this procedure. 
Note that by selecting a set of increasing sampling densities,
the SLADS algorithm can be trained to accurately predict the $ERD$ 
in conditions where the sampling density is low or high.
Intuitively, by sampling a variety of images with a variety of sampling densities, 
the final training database is constructed to represent the behaviors that will occur as the sampling density increases
when SLADS is used.

\subsection{Approximating the reduction-in-distortion}
\label{sec:Approx RD}

While the training procedure described in Section~\ref{sec:efficient training} is possible,
it is very computationally expensive because of the need to exactly compute 
the value of $R^{(s)}$.
Since this computation is done in the training phase,
we rewrite equation~(\ref{eqn:RD}) without the dependence on $k$ to be
\begin{eqnarray}
R^{(s)} = D ( X , \hat{X} ) - D ( X , \hat{X}^{(s)} ) \ ,
\label{eqn:RD2}
\end{eqnarray}
where $X$ is known in training, 
$\hat{X}$ is the reconstruction using the selected sample points,
and $\hat{X}^{(s)}$ is the reconstruction using the selected sample points plus the value $X_s$.
Notice that $\hat{X}^{(s)}$ must be recomputed for each new pixel $s$.
This requires that a full reconstruction be computed for each entry of the training database.
While this maybe possible, it typically represents a very large computational burden.

In order to reduce this computational burden,
we introduce in this section a method for approximating the value of $R^{(s)}$
so that only a single reconstruction must be performed in order to evaluate $R^{(s)}$ for all pixels $s$ in an image.
This dramatically reduces the computation required to build the training database.

In order to express our approximation,
we first rewrite the reduction-in-distortion in the form
$$
R^{(s)} = \sum_{r\in \Omega} R_r^{(s)},
$$
where
$$
R_r^{(s)} = D ( X_r , \hat{X}_r ) - D ( X_r , \hat{X}_r^{(s)} ).
$$
So here $R_r^{(s)}$ is the reduction-in-distortion at pixel $r$ 
due to making a measurement at pixel $s$.
Using this notation, our approximation is given by
\begin{equation}
R_r^{(s)} \approx \tilde{R}_r^{(s)} = h_{s,r}  D\left( X_r,\hat{X}_r \right) \ ,
\label{eqn:crucial approximation}
\end{equation}
where 
\begin{equation}
h_{s,r} =\exp \left\{ -\frac{1}{2 \sigma_s^2 } \Vert r-s \Vert^2 \right\}  
\label{eqn:kernel}
\end{equation}%
and $\sigma_s$ is the distance between the pixel $s$ and the nearest previously measured pixel divided by a user selectable parameter $c$.
More formally, $\sigma_s$ is given by
\begin{equation}
\sigma_s = \frac{\min_{ t \in \mathcal{S} } \Vert s-t \Vert }{c} ,
\label{eqn:kernel sigma}
\end{equation}
where $\mathcal{S}$
is the set of measured locations.
So this results in the final form for the approximate reduction-in-distortion given by
\begin{equation}
\tilde{R}^{(s)}  = \sum_{r\in \Omega } h_{s,r} D\left( X_r,\hat{X}_r \right) \ ,
\label{eqn:ApproximationReductionInDistortion}
\end{equation}
where $c$ is a parameter that will be estimated for the specific problem.

In order to understand the key approximation of equation~(\ref{eqn:crucial approximation}),
notice that the reduction-in-distortion is proportional to the product 
of $h_{s,r}$ and $D\left( X_r,\hat{X}_r \right)$.
Intuitively, $h_{s,r}$ represents the weighted distance of $r$ from the location 
of the new measurement, $s$;
and $D\left( X_r,\hat{X}_r \right)$ is the initial distortion at $r$ before the measurement was made. 
So for example, when $r=s$, then $h_{s,r} =1$ and we have that
\begin{equation}
\tilde{R}_s^{(s)} = D\left( X_s, \hat{X}_s\right) \ .
\end{equation}
In this case, the reduction-in-distortion is exactly the initial distortion
since the measurement is assumed to be exact.
However, as $r$ becomes more distant from the pixel being measured, $s$, the reduction-in-distortion will be attenuated by the weight $h_{s,r} <1$.

\begin{figure}
\centering
\subfigure[Measurements]{%
\includegraphics[height=0.9in]{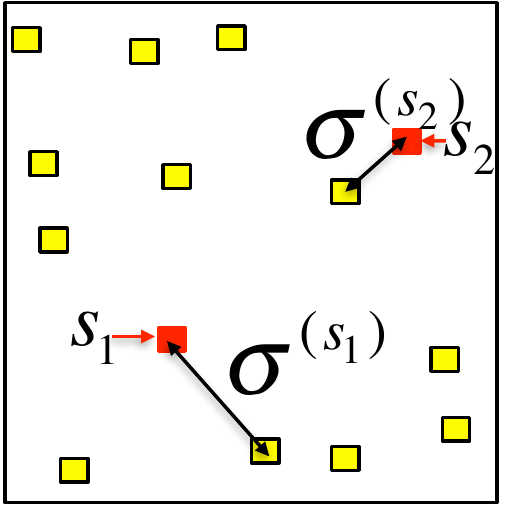}}
\quad
\subfigure[$h_{s_1,r}$]{%
\includegraphics[height=0.9in]{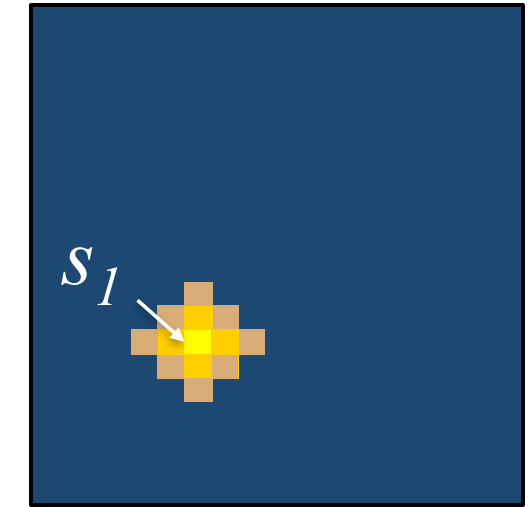}}
\quad
\subfigure[$h_{s_2,r}$]{%
\includegraphics[height=0.9in]{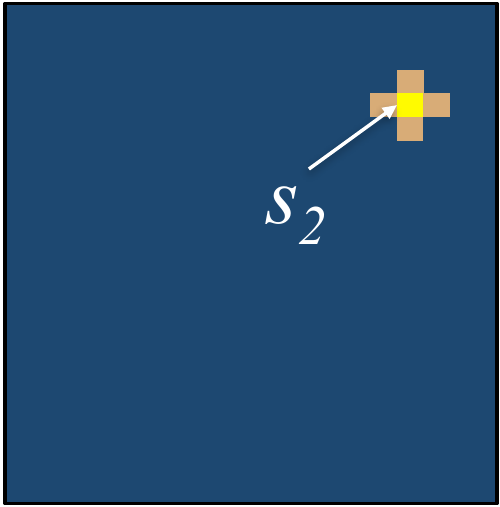}}
\caption{
This figure illustrates the shape of the function $h_{s,r}$ as a function of $r$.
(a) The red squares represent the two new measurement locations $s_1$ and $s_2$,
and the yellow squares represent the locations of previous measurements.
(b) The function $h_{s_1,r}$ resulting from a measurement at location $s_1$.
(c) The function $h_{s_2,r}$ resulting from a measurement at location $s_2$.
Notice that since $\sigma_{s_1} > \sigma_{s_2}$ then the weighting function
$h_{s_1,r}$ is wider than the function $h_{s_2,r}$.
}
\label{fig:kernel examples}
\end{figure}

Figures~\ref{fig:kernel examples}(b) and~(c) illustrate the shape of $h_{s,r}$ for two different cases.
In Figures~\ref{fig:kernel examples}(b), the pixel $s_1$ is further from the nearest measured pixel, 
and in Figures~\ref{fig:kernel examples}(c), the pixel $s_2$ is nearer.
Notice that as $r$ becomes more distant from the measurement location $s$,
 the reduction-in-distortion becomes smaller.

\subsection{Estimating the $c$ parameter}
\label{sec:selecting c}

In this section, we present a method for estimating the parameter $c$ used 
in equation~(\ref{eqn:ApproximationReductionInDistortion}).
To do this, we create a training database that contains the approximate reduction-in-distortion
for a set of parameter values.
More specifically, each entry of the training database has the form
$$
\left(\tilde{R}^{(s;c_1)},\tilde{R}^{(s;c_2)}, \hdots, \tilde{R}^{(s;c_n)}, V^{(s)} \right),
$$
where $c \in \left\lbrace c_1,c_2,\hdots c_n \right\rbrace$ is a set of $n$ possible parameter values,
and $\tilde{R}^{(s;c_i)}$ is the approximate reduction-in-distortion computed using the parameter value $c_i$.

Using this training database, we compute the $n$ associated parameter vectors $\hat{\theta}^{(c_i)}$,
and using these parameter vectors, we apply the SLADS algorithm on $M$ images
and stop each simulation when $K$ samples are taken. 
Then for each of these $M$ SLADS simulations, we compute the total distortion as
\begin{equation}
TD_k^{(m,c_i)} = \frac{1}{\vert \Omega \vert} D \left( X^{(m)},\hat{X}^{(k,m,c_i)} \right),
\end{equation}
where $X^{(m)}$ is the $m^{th}$ actual image,
and $\hat{X}^{(m,k,c_i)}$ is the associated image reconstructed using the first $k$ samples and parameter value $c_i$.
Next we compute the average total distortion over the $M$ training images given by
\begin{equation}
\overline{TD}_k^{(c_i)} = \frac{1}{M} \displaystyle\sum\limits_{m =1}^{M} TD_k^{(m,c_i)}.
\end{equation}
From this, we then compute the area under the $\overline{TD}$ curve as the overall distortion metric for each $c_i$ given by 
\begin{equation}
DM^{(c_i)} = \displaystyle\sum\limits_{k =2}^{K} \frac{\overline{TD}_{k-1}^{(c_i)}+\overline{TD}_{k}^{(c_i)}}{2} \ ,
\label{eqn:MQ}
\end{equation}
where $K$ is the total number of samples taken before stopping.
The optimal parameter value, $c^*$, is then selected to minimize the overall distortion metric given by
\begin{equation}
c^{*} =  \arg \min_{c \in \left\lbrace c_1,c_2,\hdots c_u \right\rbrace} \left\lbrace DM^{(c)} \right\rbrace.
\end{equation}

\subsection{Local Descriptors in Feature Vector $V_s$}
\label{sec:descriptors}

In our implementation, the feature vector $V_s$ is formed 
using terms constructed from the $6$ scalar descriptors $Z_{s,1},Z_{s,2},\hdots Z_{s,6}$ listed in Table~\ref{table:features}.
More specifically, we take all the unique second-degree combinations formed from these descriptors to form $V_s$.
This gives us a total of $28$ elements for the vector $V_s$. 
\begin{align*}\label{eqn:feature vector}
V_s  =
 [ & 1, \hspace{1mm} Z_{s,1}, \hspace{1mm} \hdots , \hspace{1mm} Z_{s,6}, \hspace{1mm} Z_{s,1}^2, \hspace{1mm} Z_{s,1}Z_{s,2}, \hspace{1mm} \hdots, \hspace{1mm} Z_{s,1}Z_{s,6},\\ 
	   & Z_{s,2}^2, \hspace{1mm} Z_{s,2}Z_{s,3} , \hspace{1mm} \hdots \hspace{1mm}, Z_{s,2}Z_{s,6}, \hdots \hspace{1mm}, Z_{s,6}^2 ].
\end{align*}

The first two descriptors in Table \ref{table:features} are the gradients of the image computed from the measurement in the horizontal and vertical directions. 
The second two descriptors are measures of the variance for each unmeasured location.
So the first four descriptors quantify the intensity variation at each unmeasured location. 
Then the last two descriptors quantify how densely (or sparsely) the region surrounding an unmeasured pixel is measured. 
In particular, the first of these descriptors is the distance from the nearest measurement to an unmeasured pixel 
and the second is the area fraction that is measured in a circle surrounding an unmeasured pixel.

\begin{table}   
\begin{tabular}{|p{0.45\columnwidth}|p{0.45\columnwidth}|} 
\hline
\multicolumn{2}{|c|}{ }\\
\multicolumn{2}{|c|}{\normalsize \textbf{Measures of gradients}} \\
\multicolumn{2}{|c|}{ }\\
\hline
\vspace{2mm}
\small Gradient of the reconstruction in horizontal ($x$) direction.
\small
\begin{equation*}
Z_{s,1}= D \left( \hat{X}_{ s_{x+}},\hat{X}_{ s_{x-}}\right)
\end{equation*} 
\small where, $s_{x+}$ and $s_{x-}$ are pixels adjacent to $s$ the horizontal direction 
& 
\vspace{2mm}
\small Gradient of the reconstruction in vertical ($y$) direction.
\small
\begin{equation*}
Z_{s,2}  = D \left( \hat{X}_{ s_{y+}},\hat{X}_{ s_{y-}}\right)
\end{equation*}
\small  where, $s_{x+}$ and $s_{x-}$ are pixels adjacent to $s$ the horizontal direction 
 \\ \hline
 \multicolumn{2}{|c|}{ }\\
 \multicolumn{2}{|c|}{\normalsize \textbf{Measures of standard deviation}} \\
 \multicolumn{2}{|c|}{ }\\
\hline 
\small 
\begin{equation*}
Z_{s,3} =\sqrt{ \frac{1}{L} \displaystyle\sum\limits_{r \in \partial s}D\left(X_r,\hat{X}_s \right)^2}
\end{equation*} 
\small Here $\partial s$ is the set containing the indices 
of the $L$ nearest measurements to $s$.
&
\small
\begin{equation*} 
Z_{s,4}=\displaystyle\sum\limits_{r \in \partial s} w_{r}^{(s)}D\left(X_r,\hat{X}_s\right)
\end{equation*} 
\small Here,
\small
\begin{equation*} 
w_{r}^{(s)} =\frac{\frac{1}{\Vert s-r \Vert^2}}{\displaystyle\sum\limits_{u \in \partial s} \frac{1}{\Vert s-u \Vert^2} }
\end{equation*} 
\small and $\Vert s-r \Vert$ is the euclidean distance between $s$ and $r$.\\
\hline
\multicolumn{2}{|c|}{ }\\
 \multicolumn{2}{|c|}{\normalsize \textbf{Measures of density of measurements}} \\
 \multicolumn{2}{|c|}{ }\\
\hline 
\small
\begin{equation*} 
Z_{s,5}=\underset{r \in \partial s} {min} \Vert s-r \Vert_2
\end{equation*} \normalsize The distance from $s$ to the closest measurement.
&
\small
\begin{equation*}
Z_{s,6} = \frac{1+A_{(s;\lambda)}}{1+A_{(s;\lambda)}^{*}}
\end{equation*} 
\small Here $A_{(s;\lambda)}$ is the area of a circle $\lambda\%$ the size of the image. 
$A_{(s;\lambda)}^{*}$ is the measured area inside $A_{(s;\lambda)}$. \\
\hline
\end{tabular}
\vspace{3mm}
\caption{ List of descriptors 
used to construct the feature vector $V_s$. 
There are three main categories of descriptors: 
measures of gradients, 
measures of standard deviation, 
and measures of density of measurements surrounding the pixel $s$.}
\label{table:features}
\end{table}


\begin{figure}[h!]
\centering
\includegraphics[height=2.6in]{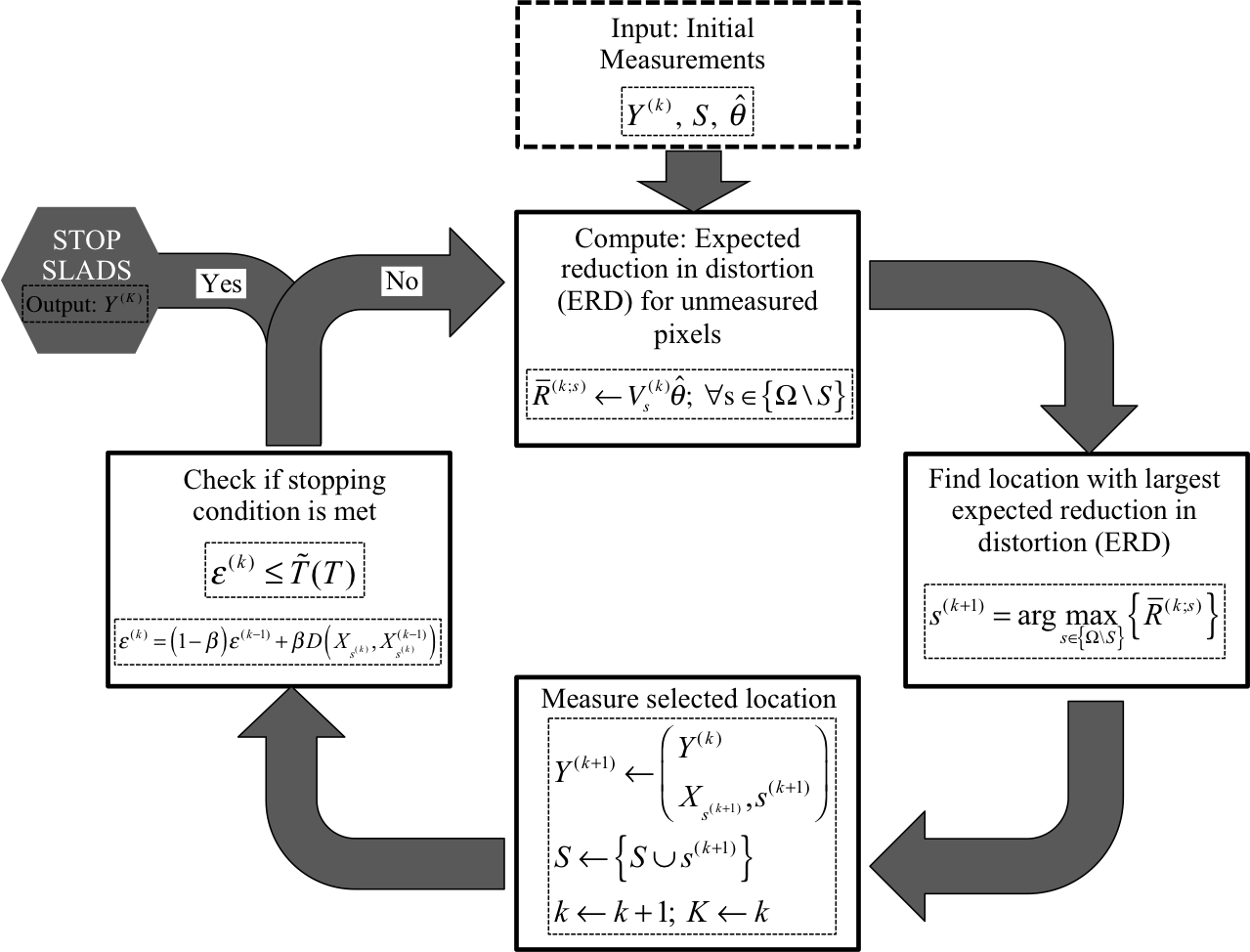}
\caption{The SLADS algorithm as a flow diagram. The inputs to the function are 
the initial measurements $Y^{(k)}$, 
the coefficients needed to compute the $ERD$, found in training, $\hat{\theta}$,
and the set $\mathcal{S} $ containing the indices of the measurements.
When the stopping condition is met 
the function will output the selected set of measurements $Y^{(K)}$.}
\label{fig:flow diagram SLADS}
\end{figure}

\section{Stopping Condition for SLADS}
\label{sec:stopping condition}

In applications, it is often important to have a stopping criteria for terminating the SLADS sampling process.
In order to define such a criteria, we first define the expected total distortion (ETD) at step $k$ by
$$
\ETD_k = \mathbb{E} \left[ \frac{1}{\vert \Omega \vert} D\left( X ,\hat{X}^{(k)} \right) \right] \ .
\label{eqn: ETD_k}
$$
Notice that the expectation is necessary since the true value of $X$ is unavailable during the sampling process.
Then our goal is to stop sampling when the ETD falls below a predetermined threshold.
$$
\ETD_k \leq  T 
\label{eqn: ETD_k threshold}
$$

In order to make this practical,
at each step of the SLADS algorithm we compute
\begin{equation}
\epsilon^{(k)} = (1-\beta) \epsilon^{(k-1)} +\beta D\left( X_{s^{(k)}} , \hat{X}_{s^{(k)}}^{(k-1)} \right),
\label{eqn:stopping condition}
\end{equation}
where $k>1$, $\beta$ is a user selected parameter that determines the amount
of temporal smoothing,
$X_{s^{(k)}}$ is the measured value of the pixel at step $k$,
and $\hat{X}_{s^{(k)}}^{(k-1)}$ is the reconstructed value of the same pixel at step $k-1$.

Intuitively, the value of $\epsilon^{(k)}$ measures the average 
level of distortion in the measurements.
So a large value of $\epsilon^{(k)}$ indicates that more samples
need to be taken, and a smaller value indicates that the reconstruction
is accurate and the sampling process can be terminated.
However, in typical situations, it will be the case that
$$
\epsilon^{(k)} > \ETD_k
$$
because the SLADS algorithm will tend to select pixels to measure whose values have great uncertainty.

In order to compensate for this effect, we compute a function $\tilde{T} (T)$ using a look-up-table (LUT)
and stop sampling when
$$
\epsilon^{(k)} \leq \tilde{T} (T).
$$
The function $\tilde{T} (T)$ is determined using a set of training images, $\{ X_1, \cdots , X_M\}$.
For each image, we first determine the number of steps, $K_m(T)$, required to achieve the desired distortion.
\begin{equation}
K_m (T)=\min_{k} \left\lbrace k: \frac{1}{\vert \Omega \vert} D\left( X ,\hat{X}^{(k)} \right) \leq T \right\rbrace.
\label{eqn:find all T from stop DB}
\end{equation}
Then we average the value of $\epsilon^{k}$ for each of the $M$ images to determine the adjusted threshold:
\begin{equation}
\tilde{T} (T) = \frac{1}{M}\displaystyle\sum\limits_{m =1}^{M} \epsilon_m^{(K_{m}(T))}.
\label{eqn:find avg T from stop DB}
\end{equation}

Selection of the parameter $\beta$ allows the estimate of $\epsilon^{k}$ to be smoothed as a function of $k$.
In practice, we have used the following formula to set $\beta$:
\begin{equation*}
\beta =\begin{cases}
 0.001 \left(\frac{ \log_2 \left(512^2\right) -  \log_2 \left(  \vert \Omega \vert  \right)}{2} +1 \right) \quad  \vert \Omega \vert \leq 512^2 \\
0.001 \left(\frac{ \log_2 \left( \vert \Omega \vert \right) -  \log_2 \left( 512^2 \right)}{2} +1 \right)^{-1} \quad \vert \Omega \vert > 512^2
\end{cases}
\end{equation*}
where $\vert \Omega \vert$ is the number of pixels in the image.

Figure~\ref{fig:flow diagram SLADS} shows the SLADS algorithm as a flow diagram after the stopping condition in the previews section is incorporated.

\section{Group-wise SLADS}
\label{sec:group-wise SLADS}

\begin{figure*}
\centering
\fbox{
 \addtolength{\linewidth}{-6\fboxsep}%
 \addtolength{\linewidth}{-6\fboxrule}%
 \begin{minipage}{\linewidth}
\begin{algorithmic}
{
\Function{$Y^{(K)}$ $\gets$ \textsc{Group-wise SLADS}}{$Y^{(k)},\hat{\theta},k,B$}
\State \tiny
\State \normalsize $\mathcal{S} \gets \left\lbrace s^{(1)},s^{(1)},\hdots s^{(k)} \right\rbrace $
\State \tiny
\normalsize  \While  {Stopping condition not met} 
\State \tiny
\normalsize \For{\ $b =1,\hdots B$} 
\State \tiny \normalsize
\State Form \textit{pseudo}-measurement vector $Y^{(k+1)}_{b}$ as shown in equation (\ref{eqn:Group Sampling Update Measurement Vector})
\State \tiny \normalsize
\State Compute \textit{pseudo}-$ERD$ $\bar{R}^{(k,s^{(k+1)}_1,s^{(k+1)}_2 \hdots s^{(k+1)}_{b-1} ;s)}$ from $Y^{(k+1)}_{b}$ $\forall s \in \mathbb{S}$ using equation (\ref{eqn:Group Sampling Compute ERD})
\State \tiny \normalsize
\State $s^{(k+1)}_{b} =\arg \underset{s \in  \left\lbrace \Omega \setminus \mathcal{S} \right\rbrace }{\operatorname{ max}} \left( \bar{R}^{(k,s^{(k+1)}_1,s^{(k+1)}_2 \hdots s^{(k+1)}_{b-1} ;s)} \right)$ 
\State \tiny \normalsize
\State $\mathcal{S} \gets \left\lbrace \mathcal{S} \cup s^{(k+1)}_b \right\rbrace$
\State \tiny \normalsize
\EndFor
\State \tiny \normalsize
$Y^{(k+1)} = 
\left[ 
\begin{array}{c}
Y^{(k)}\\
s^{(k+1)}_1, \hat{X}_{s^{(k+1)}_1}^{(k)} \\
s^{(k+1)}_2, \hat{X}_{s^{(k+1)}_2}^{(k)}\\
\vdots \\
s^{(k+1)}_b, \hat{X}_{s^{(k+1)}_b}^{(k)}
\end{array}
\right] \ $
\State
\State \scriptsize
\State \normalsize $k\leftarrow k+1$
\State \tiny
\normalsize  \EndWhile \normalsize
\State \normalsize \tiny
\State \normalsize $K\gets k$
\State \tiny
\normalsize  \EndFunction}
\end{algorithmic}
 \end{minipage}
}
\vspace{0.5mm}
\caption{Our proposed method for Group-wise SLADS. Here instead of just selecting $1$ measurement in each step of SLADS, as we did in Figure~\ref{fig:SLADS pseudocode}, we now select $B$ measurement locations at each step.}  
\label{fig:Group SLADS pseudocode}
\end{figure*}

In this section, we introduce a group-wise SLADS approach
in which $B$ measurements are made at each step of the algorithm.
Group-wise SLADS may be more appropriate in applications
where it is faster to measure a set of predetermined pixels in a single burst.
So at the $k^{th}$ step, our goal will be to select a group of measurement positions, 
$\mathcal{S}^{(k+1)}=\left\lbrace s^{(k+1)}_1, s^{(k+1)}_2, \hdots s^{(k+1)}_B \right\rbrace$, that will yield the greatest expected reduction-in-distortion.
\begin{equation}
\mathcal{S}^{(k+1)} =  \arg \max_{ \left\lbrace s_1,s_2, \hdots s_B \right\rbrace \in \left\lbrace \Omega \setminus \mathcal{S} \right\rbrace } \left\lbrace \bar{R}^{\left( k;s_1,s_2,\hdots s_B \right)}  \right\rbrace,
\end{equation}
where $\bar{R}^{\left( k;s_1,s_2,\hdots s_B \right)}$ is the expected reduction-in-distortion due to measurements $s^{(k+1)}_1, s^{(k+1)}_2, \hdots s^{(k+1)}_B$. 
However, solving this problem requires that we consider ${{N- \vert \mathcal{S} \vert}\choose{B}}$ different combinations of measurements.

In order to address this problem, we introduce a method in which we choose the measurements sequentially, just as we do in standard SLADS.
Since group-wise SLADS requires that we make measurements in groups,
we cannot make the associated measurement after each location is selected.
Consequently, we cannot recompute the $ERD$ function after each location is selected,
and therefore, we cannot select the best position for the next measurement.
Our solution to this problem is to estimate the value at each selected location, 
and then we use the estimated value as if it were the true measured value.

More specifically, we first determine measurement location $s^{(k+1)}_1$ using equation (\ref{eqn:ideal sampling strategy}),
and then let $\mathcal{S} \gets \left\lbrace \mathcal{S} \cup s^{(k+1)}_1  \right\rbrace$.
Now without measuring the pixel at $s^{(k+1)}_1$,
we would like to find the location of the next pixel to measure, $s^{(k+1)}_2$.
However, since $s^{(k+1)}_1$ has now been chosen, it is important to incorporate this information when choosing the next location $s^{(k+1)}_2$. 
In our implementation, we temporarily assume that the true value of the pixel, $X_{s^{(k+1)}_1}$,
is given by its estimated value, $\hat{X}_{s^{(k+1)}_1}^{(k)}$, computed using all the measurements acquired up until the $k^{th}$ step.
We will refer to $\hat{X}_{s^{(k+1)}_1}^{(k)}$ as a \textit{pseudo}-measurement since it takes the place of a true measurement of the pixel.
Now using this \textit{pseudo}-measurement along with all previous real measurements,
we estimate a \textit{pseudo}-$ERD$ $\bar{R}^{(k,s^{(k+1)}_1;s)}$ for all $s \in \left\lbrace \Omega \setminus \mathcal{S} \right\rbrace $ and from that select the next location to measure.
We repeat this procedure to find all $B$ measurements.

So the procedure to find the $b^{th}$ measurement is as follows. 
We first construct a \textit{pseudo}-measurement vector,
\begin{equation}
Y^{(k+1)}_{b} = 
\left[ 
\begin{array}{c}
Y^{(k)} \\
s^{(k+1)}_1, \hat{X}_{s^{(k+1)}_1}^{(k)}\\
s^{(k+1)}_2, \hat{X}_{s^{(k+1)}_2}^{(k)}\\
\hdots\\
s^{(k+1)}_{b-1}, \hat{X}_{s^{(k+1)}_{b-1}}^{(k)}\\
\end{array}
\right], \ 
\label{eqn:Group Sampling Update Measurement Vector}
\end{equation}
where $Y^{(k+1)}_{1}=Y^{(k)}$.
Then using this \textit{pseudo}-measurement vector we compute the \textit{pseudo}-$ERD$ for all $s \in \left\lbrace \Omega \setminus \mathcal{S} \right\rbrace $
\begin{equation}
\bar{R}^{(k,s^{(k+1)}_1,s^{(k+1)}_2,\hdots s^{(k+1)}_{b-1};s)} = V^{(k,s^{(k+1)}_1,s^{(k+1)}_2,\hdots s^{(k+1)}_{b-1})}_{s} \hat{\theta}.
\label{eqn:Group Sampling Compute ERD}
\end{equation}
where $V^{(k,s^{(k+1)}_1,s^{(k+1)}_2,\hdots s^{(k+1)}_{b-1})}_{s}$ is the feature vector that corresponds to location $s$.
It is important to note that when $b=1$ the \textit{pseudo}-$ERD$ is the actual $ERD$ computed using the actual measurements only.
Now we find the location that maximizes the \textit{pseudo}-$ERD$ by
\begin{equation}
s^{(k+1)}_b= \arg \max_{s \in \left\lbrace \Omega \setminus \mathcal{S} \right\rbrace } \left\lbrace \bar{R}^{(k,s^{(k+1)}_1,s^{(k+1)}_2,\hdots s^{(k+1)}_{b-1};s)} \right\rbrace.
\label{eqn:Group Sampling find next location}
\end{equation}
Then finally we update the set of measured locations by
\begin{equation}
\mathcal{S}\gets \left\lbrace S \cup s^{(k+1)}_b \right\rbrace.
\label{eqn:Group Sampling update location vector}
\end{equation}

Figure~\ref{fig:Group SLADS pseudocode} shows a detailed illustration of the proposed group-wise SLADS method.

\section{Results}
\label{sec:results}

In the following sections,
we first validate the approximation to the $ERD$,
and then we compare SLADS to alternative sampling approaches based 
on both real and simulated data.
We then evaluate the stopping condition presented in Section \ref{sec:stopping condition}
and finally compare the group-wise SLADS method presented in Section \ref{sec:group-wise SLADS} with SLADS.
The distortion metrics
and the reconstruction methods 
we used in these experiments
are detailed in Appendices~\ref{sec: distortion metrics} and~\ref{sec: Reconstruction methods}.
It is also important to note that we start all experiments 
by first acquiring $1\%$ of the image according to low-discrepancy sampling\cite{LDSampling}.

\begin{figure*}
\centering
\subfigure[{ \scriptsize}]{\includegraphics[height=1.1in,width=3in]{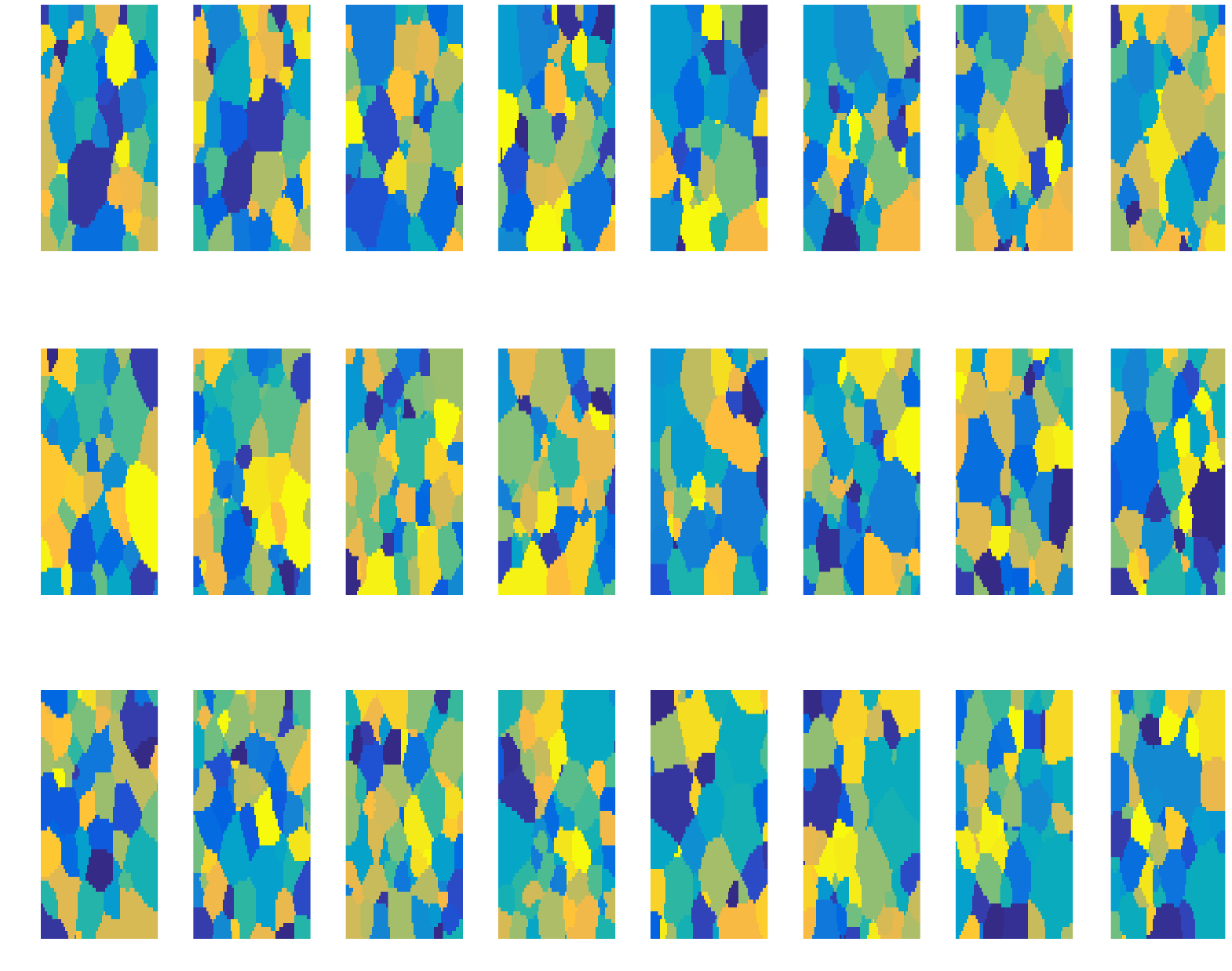}} 
 \hspace{7mm} \subfigure[{\scriptsize}]{\includegraphics[height=1.1in,width=3in]{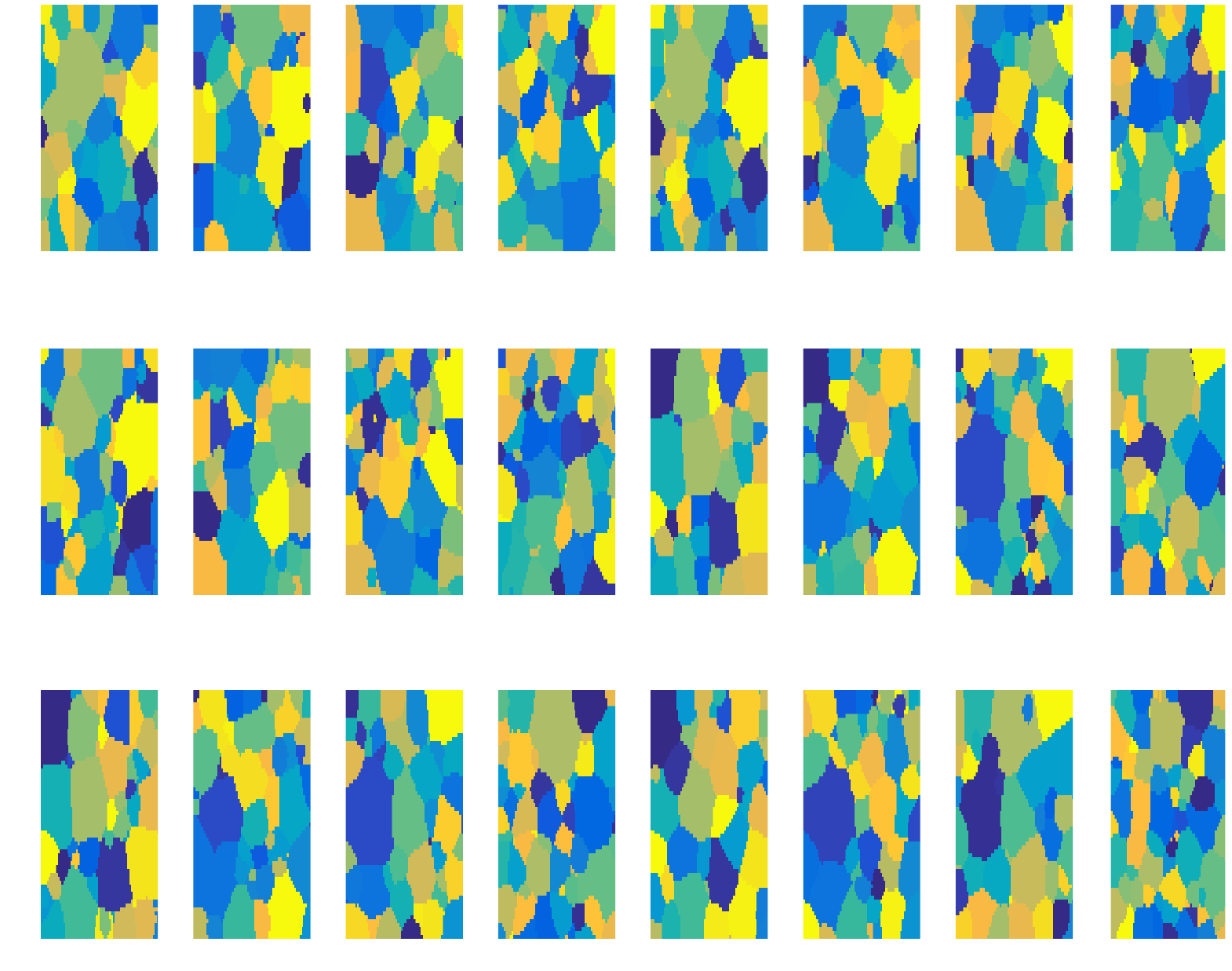}}
\caption{Images used to compute regression coefficients and perform SLADS in the experiment detailed in Section~\ref{sec:empirical validation}.
These images have $64 \times 64$ pixels each,
and are synthetic EBSD image
generated using the Dream.3D software.
The images here are discretely labeled and the different colors correspond to different crystal orientations.
In particular, (a) shows the images that were used to find the regression coefficients and 
(b) shows the images on which SLADS was performed.
}
\label{fig:ARDvRD images}
\end{figure*}
\subsection{Validating the $ERD$ Approximation}
\label{sec:empirical validation}

In this section, we compare the results using the true and approximate $ERD$ described in Section~\ref{sec:Approx RD}
in order to validate the efficacy of the approximation.
The SLADS algorithm was trained and then tested using the synthetic EBSD images shown 
in Figures~\ref{fig:ARDvRD images}(a) and \ref{fig:ARDvRD images}(b).
Both sets of images were generated using the Dream.3D software \cite{Dream3D}.

The training images were constructed to have a small size of $64 \times 64$
so that it would be tractable to compute the true reduction-in-distortion from equation~(\ref{eqn:RD2})
along with the associated true regression parameter vector $\hat{\theta}^{true}$.
This allowed us to compute the true $ERD$ for this relatively small problem.

\begin{figure}
\includegraphics[height=2in,width=3in]{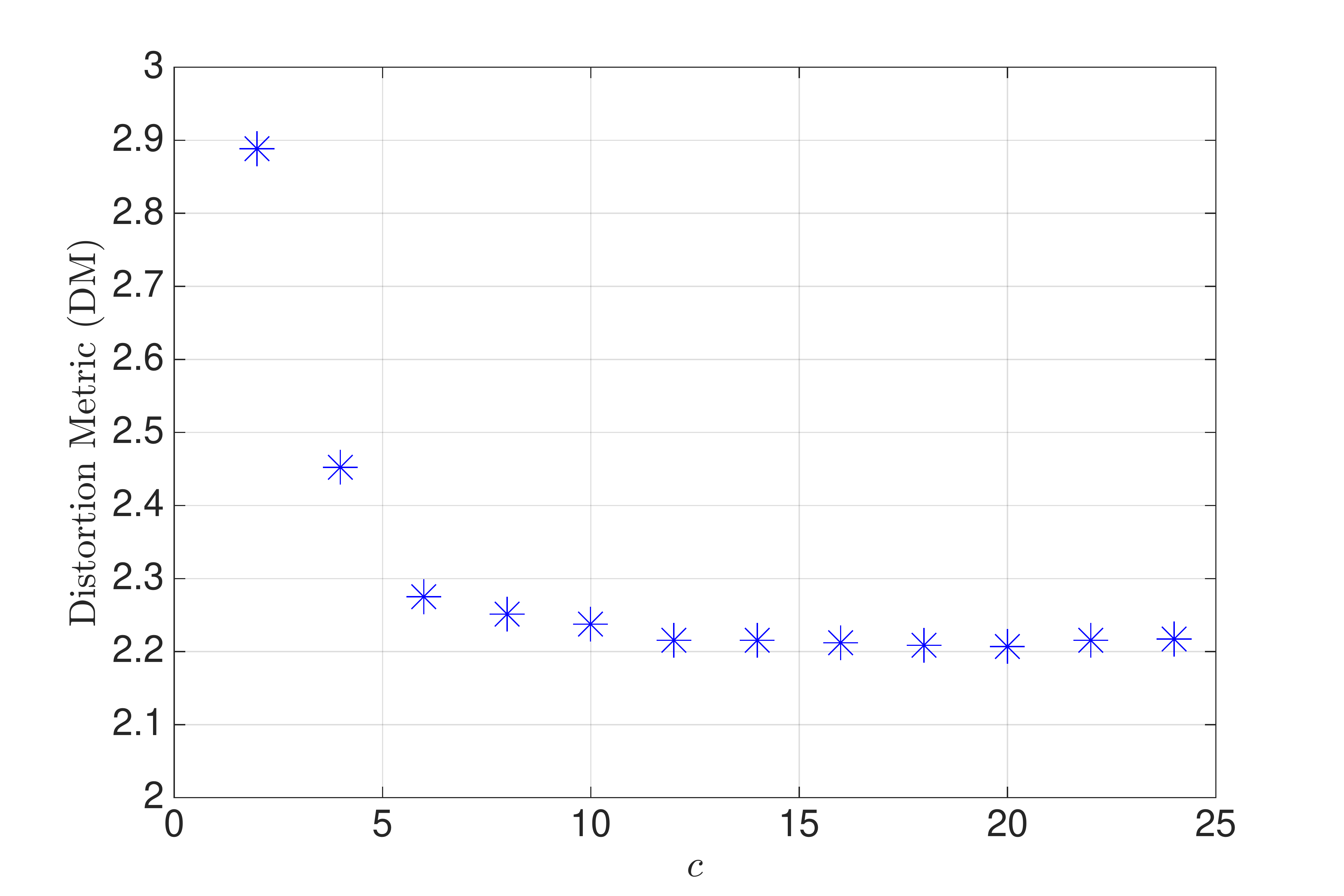}
\caption{
Experimental results from the experiment we performed 
to pick the best value for $c$ in equation~\ref{eqn:kernel sigma}.
The figure shows the overall distortion metric of equation~\ref{eqn:MQ} 
computed from the $24$ SLADS experiments for the $c=2,4,6,\hdots 24$. 
The distortion metric is smallest when $c=20$.
}
\label{fig:Pick c_i}
\end{figure}

We selected the optimal parameter value, $c^{*}$, 
using the method described in Section~\ref{sec:selecting c}
from the possible values $c\in \left\lbrace 2,4,6, \hdots 24 \right\rbrace$.
Figure~\ref{fig:Pick c_i} shows a plot of $DM^{(c_i)}$ versus $c_i$.
In this case, the optimal parameter value that minimizes the overall distortion metric is given by $c^*=20$.
However, we also note that the metric is low for a wide range of values.

\begin{figure}
\includegraphics[height=2in,width=3in]{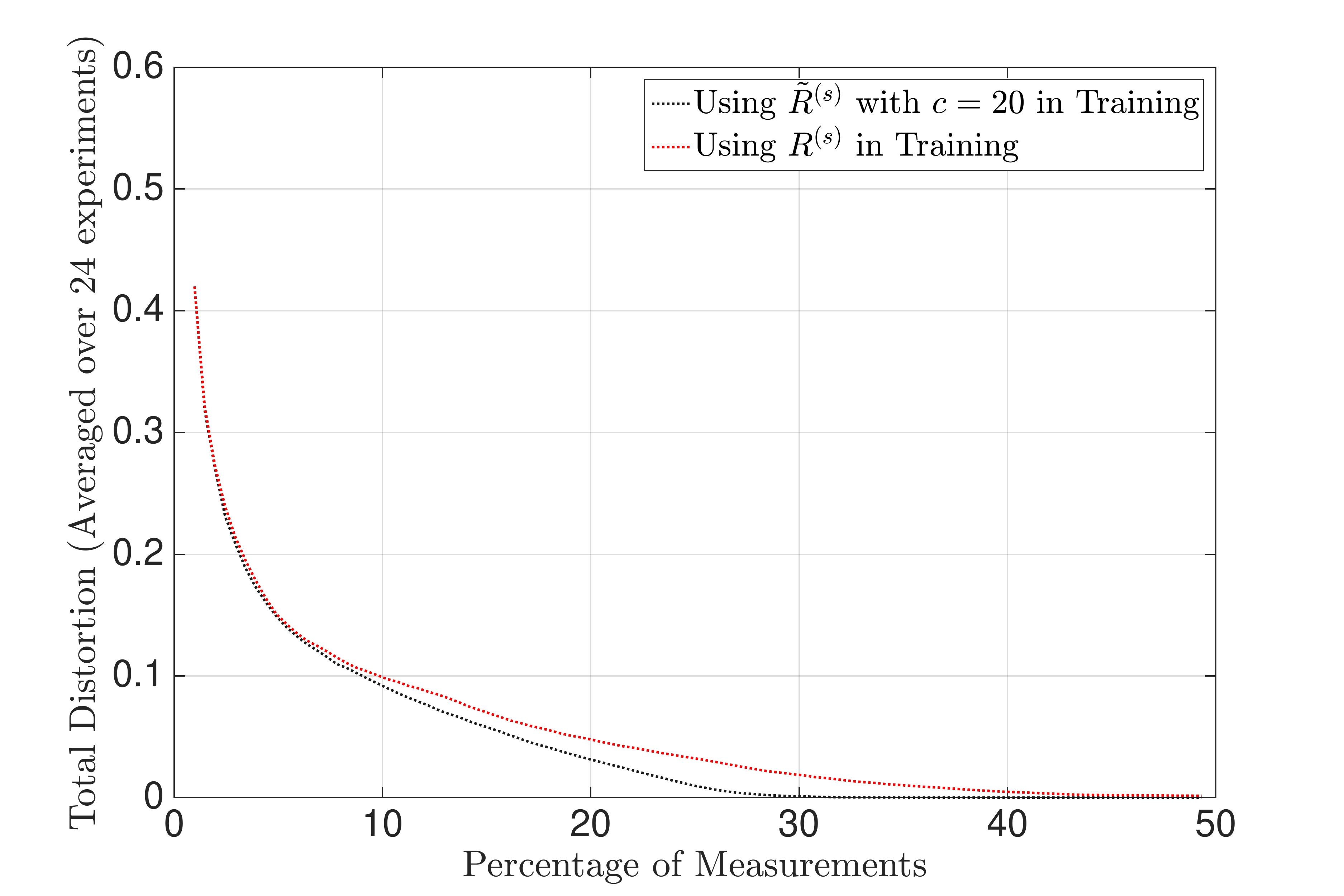}
\caption{
Experimental results 
from the experiments we performed 
to empirically validate the approximation to the $ERD$.
The curves show the TD averaged over $24$ different experiments 
versus the percentage of samples 
when the true value of $R^{(s)}$
was used to estimate $\theta$ and  
when the approximate value of $R^{(s)}$
was used to estimate $\theta$. 
In this experiments we selected $c=20$ to compute the approximate RD.}
\label{fig:ARD vs RD results}
\end{figure}

Figure~\ref{fig:ARD vs RD results} shows the results of plotting the total distortion, $\overline{TD}_k$, 
versus the percentage of samples for both the true regression parameter,
$\hat{\theta}^{true}$, and the approximate regression parameter, $\hat{\theta}^{(c^*)}$.
While the two curves are close, the approximate reduction-in-distortion results in a lower curve 
than the true reduction-in-distortion.
This indicates that the approximation is effective,
but it is surprising that the approximate parameter works better than the true parameter.
We conjecture that this suggests that there approximations in our models 
that might allow for future improvements.

\begin{figure*}
\centering
\subfigure[]{\includegraphics[height=1.8in,width=3.25in]{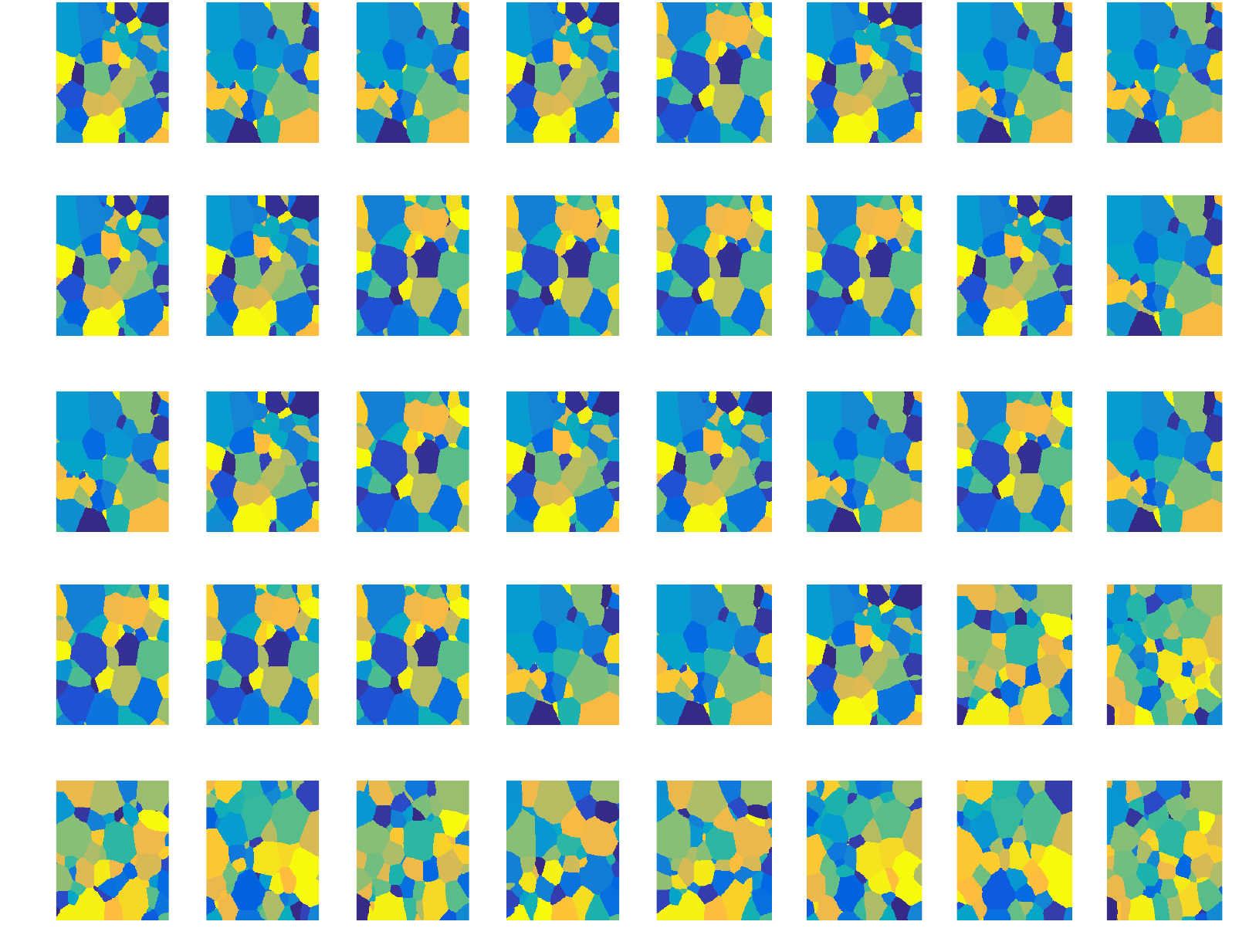}} 
\hspace{15mm} \subfigure[]{\includegraphics[height=1.8in,width=1.5in]{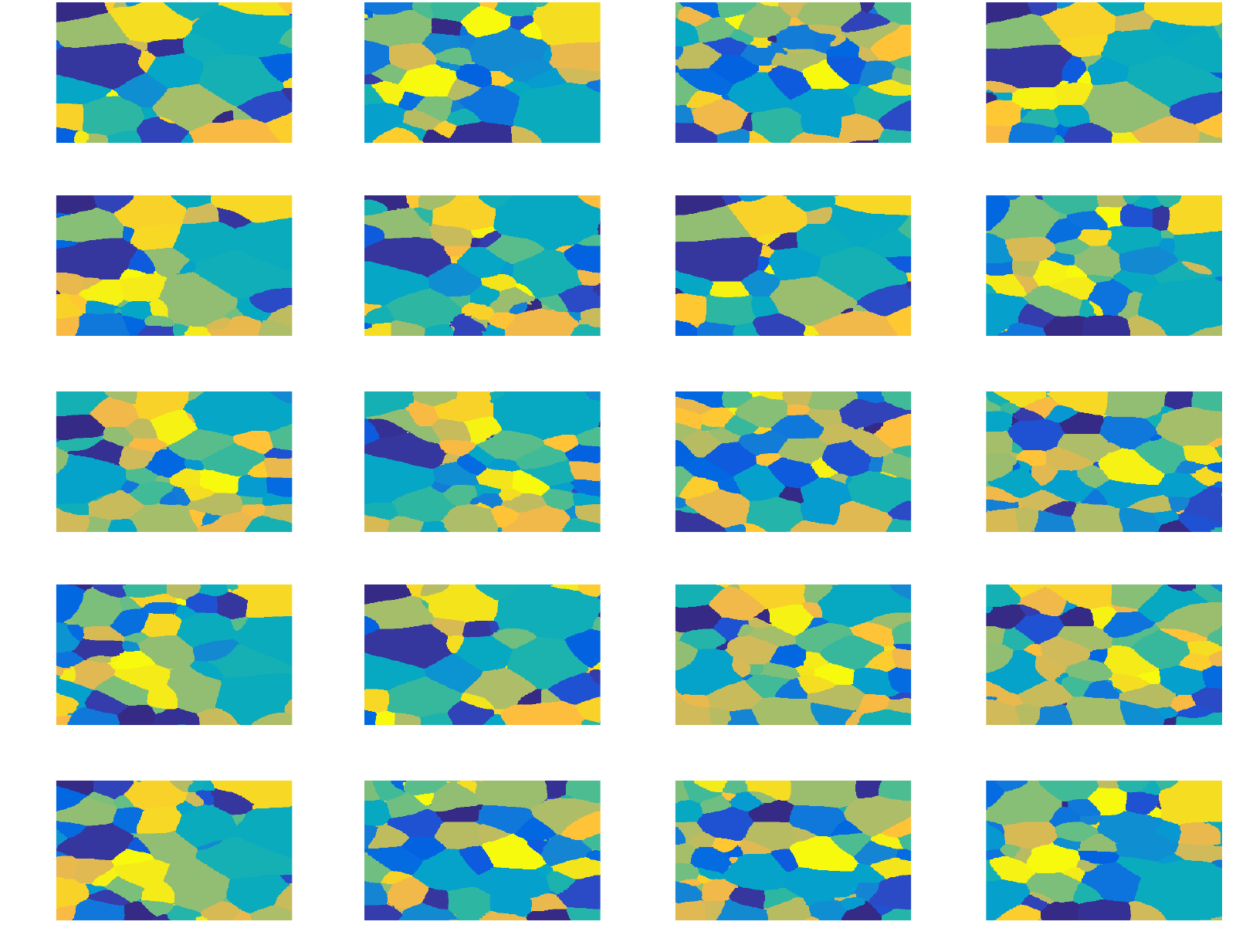}}
\caption{Images used for training and testing in the experiment detailed in Section \ref{sec:SimulatedEBSDExperiments} in which we compared SLADS with LS and RS. 
These images have $512 \times 512$ pixels each,
and are synthetic EBSD image generated using the Dream.3D software.
The images here are discretely labeled and the different colors correspond to different crystal orientations.
In particular, (a): shows the images that were used for training and 
(b): shows the images that were used for testing.
}
\label{fig:discrete images}
\end{figure*}

\begin{figure}
\centering
\includegraphics[height=2.3in,width=3in]{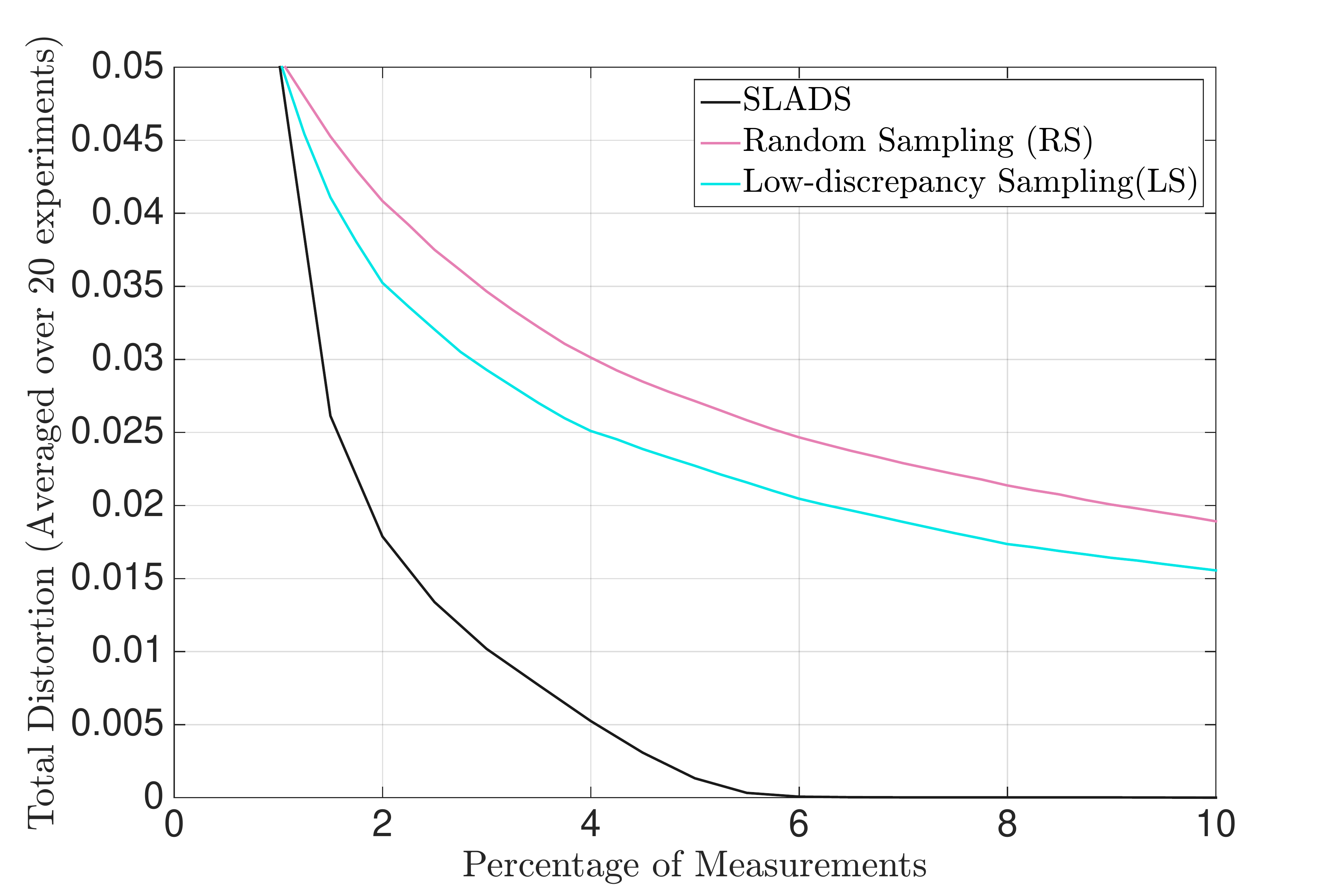}
\caption{In this figure we plot the average $TD$, $\bar{TD}_k$, (averaged over $20$ sampling experiments) 
versus the percentage of samples 
for the experiment detailed in Section~\ref{sec:SimulatedEBSDExperiments}.
The plot shows the $\bar{TD}_k$ curves 
that correspond to SLADS, RS and LS.
The images that were sampled in this experiment are shown in Figure~\ref{fig:discrete images}(b). 
}
\label{fig:discrete TD plot}
\end{figure}

\begin{figure*}
\centering
\subcapraggedrighttrue
\subfigure[{ \scriptsize Original Image}]{\includegraphics[height=1.3in,width=1.4in]{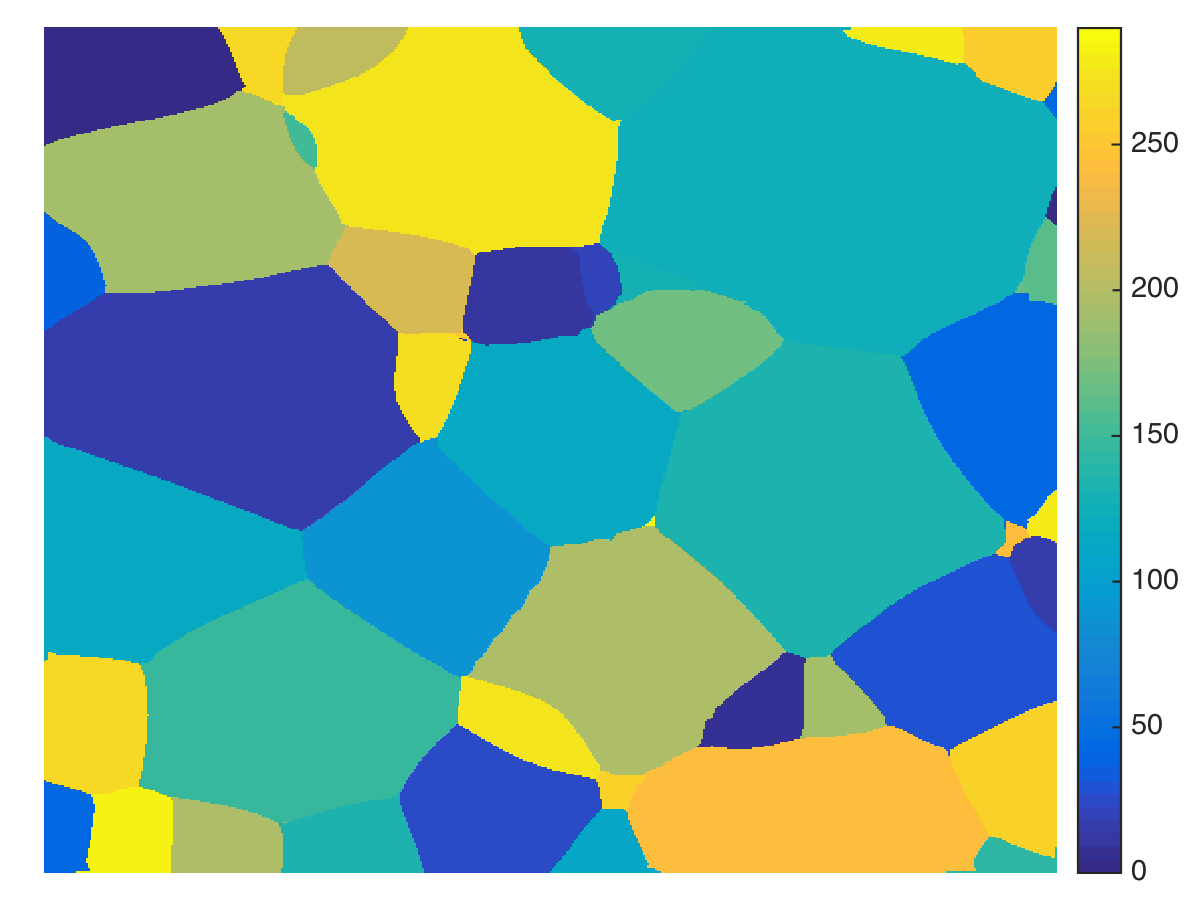}} 
\subfigure[{ \scriptsize RS: Sample locations ($\sim 6 \%$) }]
{\includegraphics[height=1.3in,width=1.4in
]{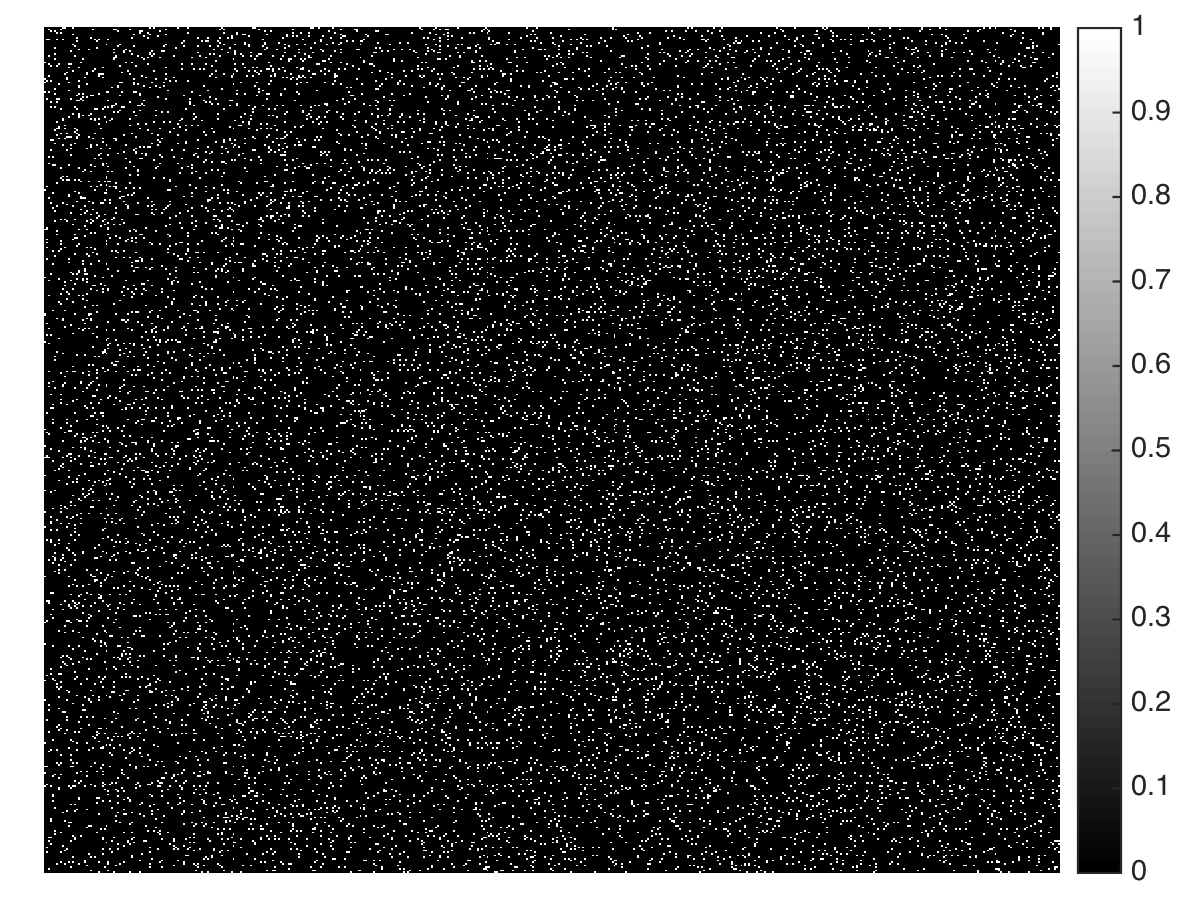}}
 \subfigure[{ \scriptsize LS: Sample locations ($ \sim  6 \%$) }]{\includegraphics[height=1.3in,width=1.4in
]{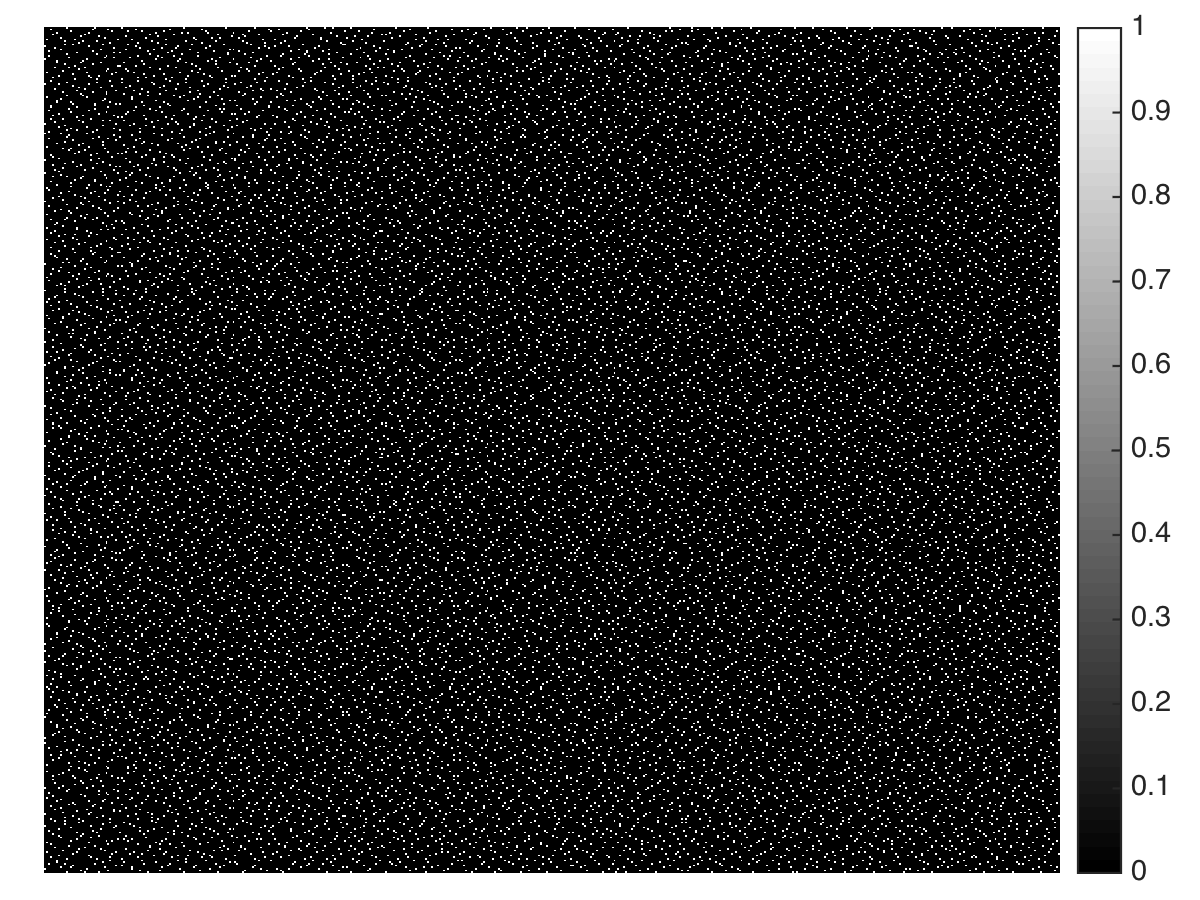}}
 \subfigure[{ \scriptsize SLADS: Sample locations ($ \sim 6 \%$) }]{\includegraphics[height=1.3in,width=1.4in
]{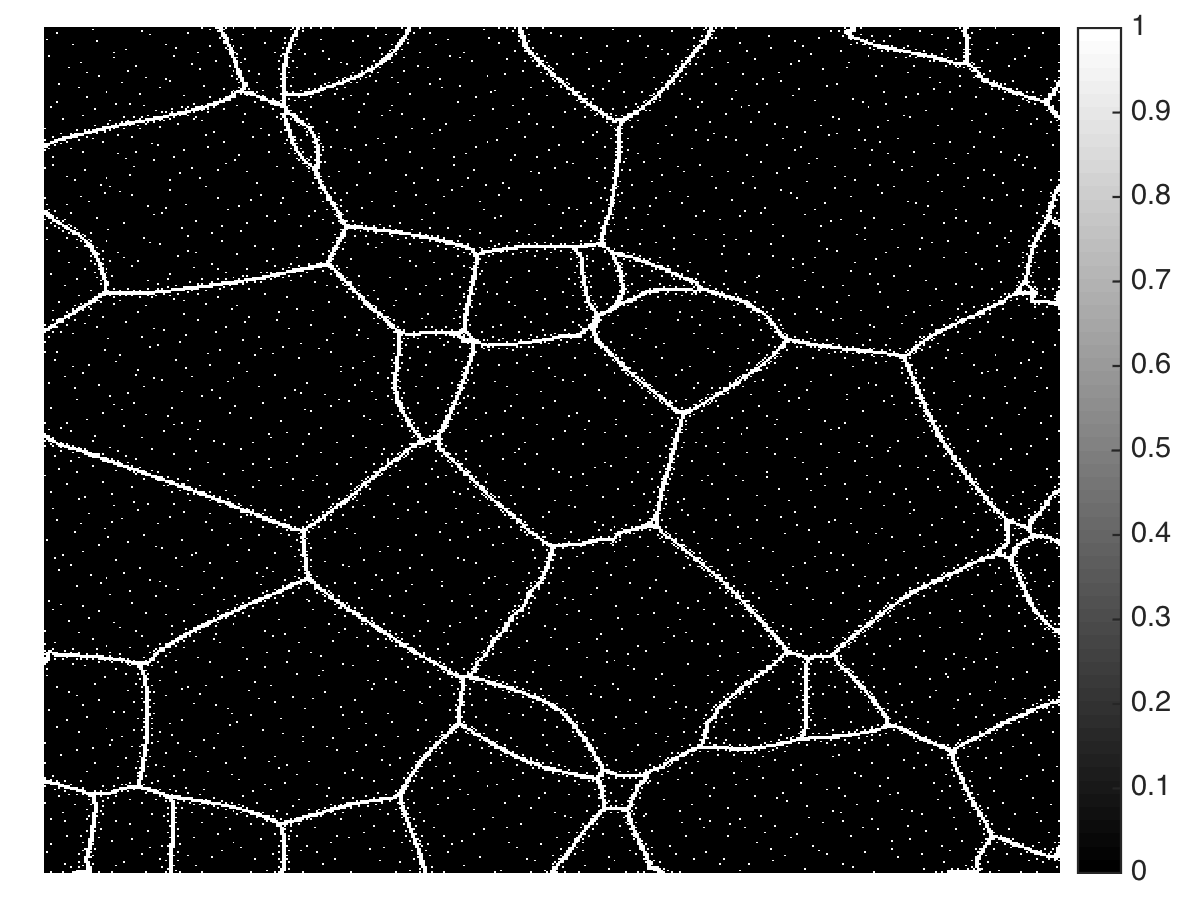}}\\

\hspace*{1.45in} \subfigure[{ \scriptsize RS: Reconstructed Image}]{\includegraphics[height=1.3in,width=1.4in]{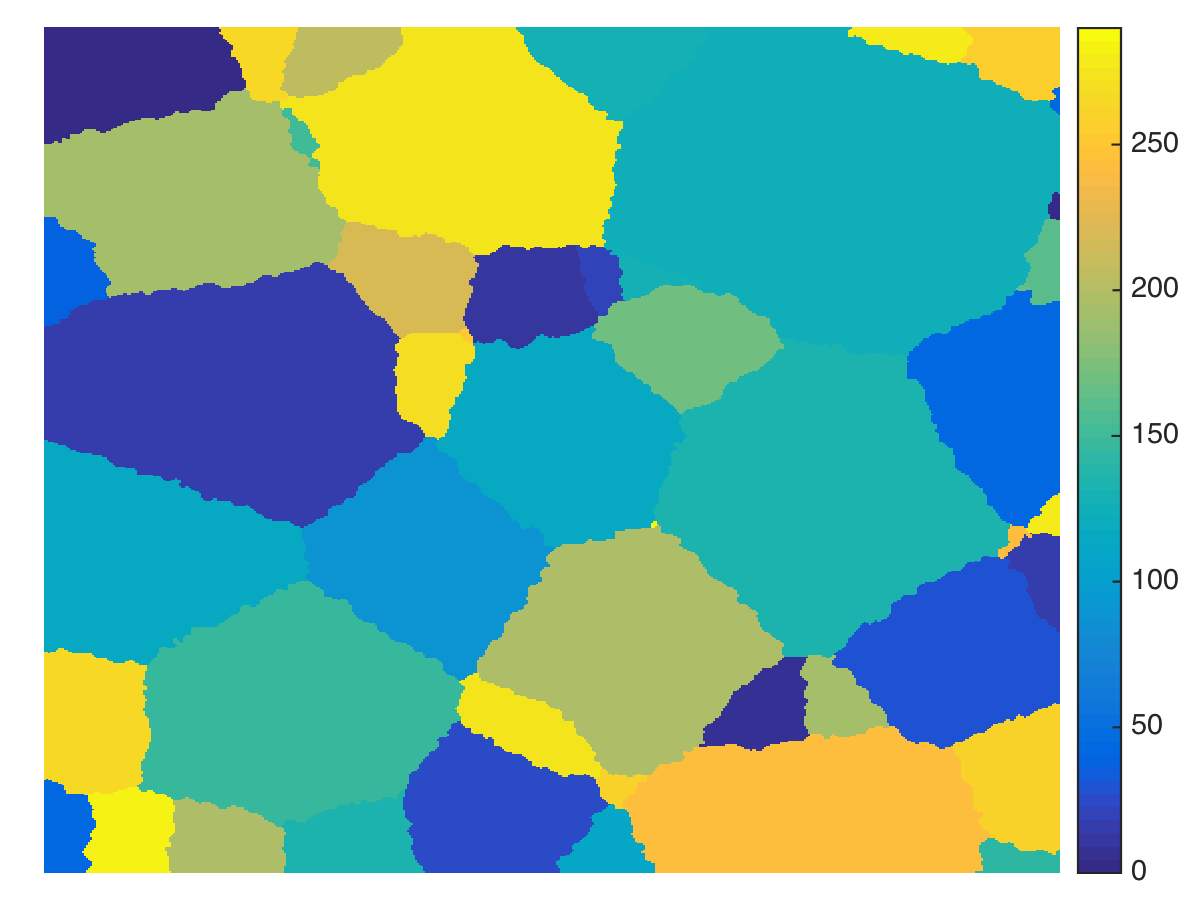}}
 \subfigure[{ \scriptsize LS: Reconstructed Image}]{\includegraphics[height=1.3in,width=1.4in]{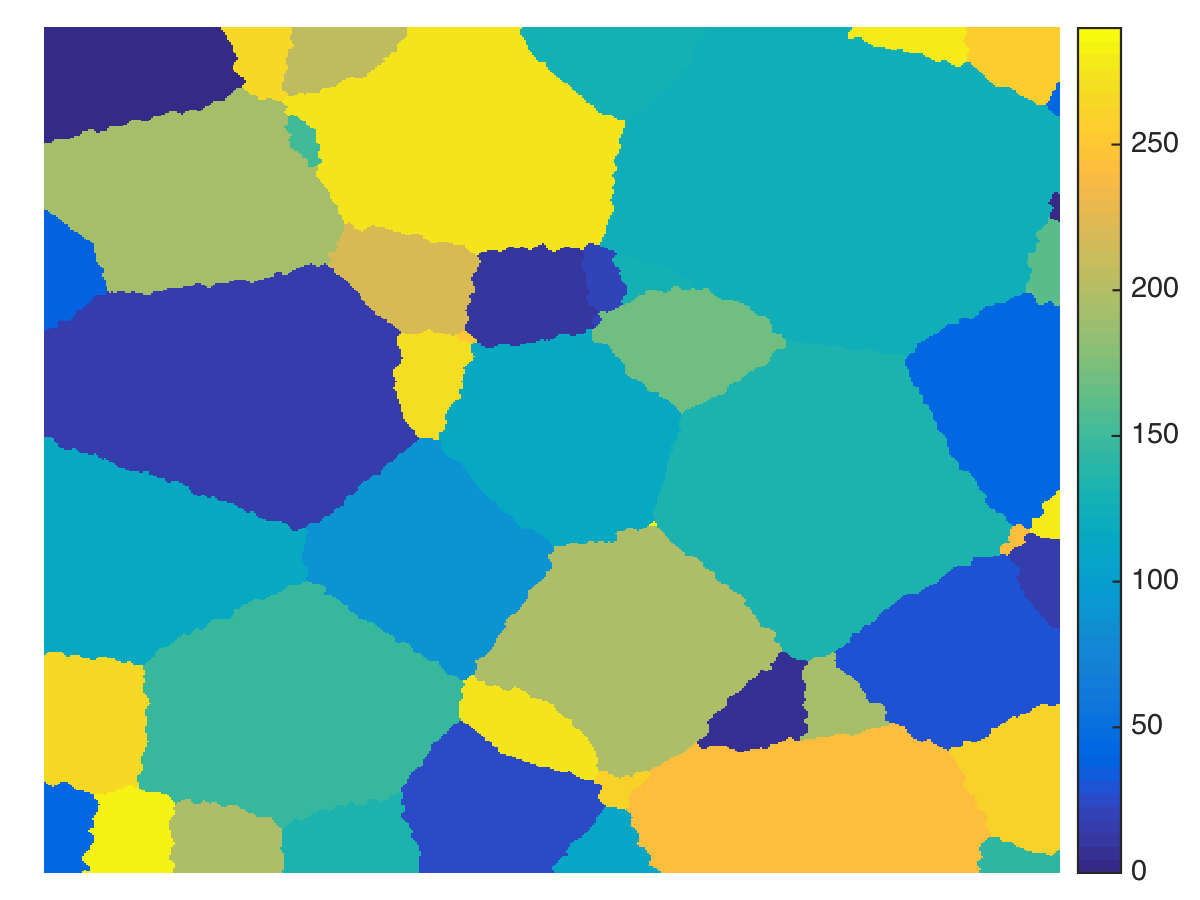}}
 \subfigure[{ \scriptsize SLADS: Reconstructed Image}]{\includegraphics[height=1.3in,width=1.4in
]{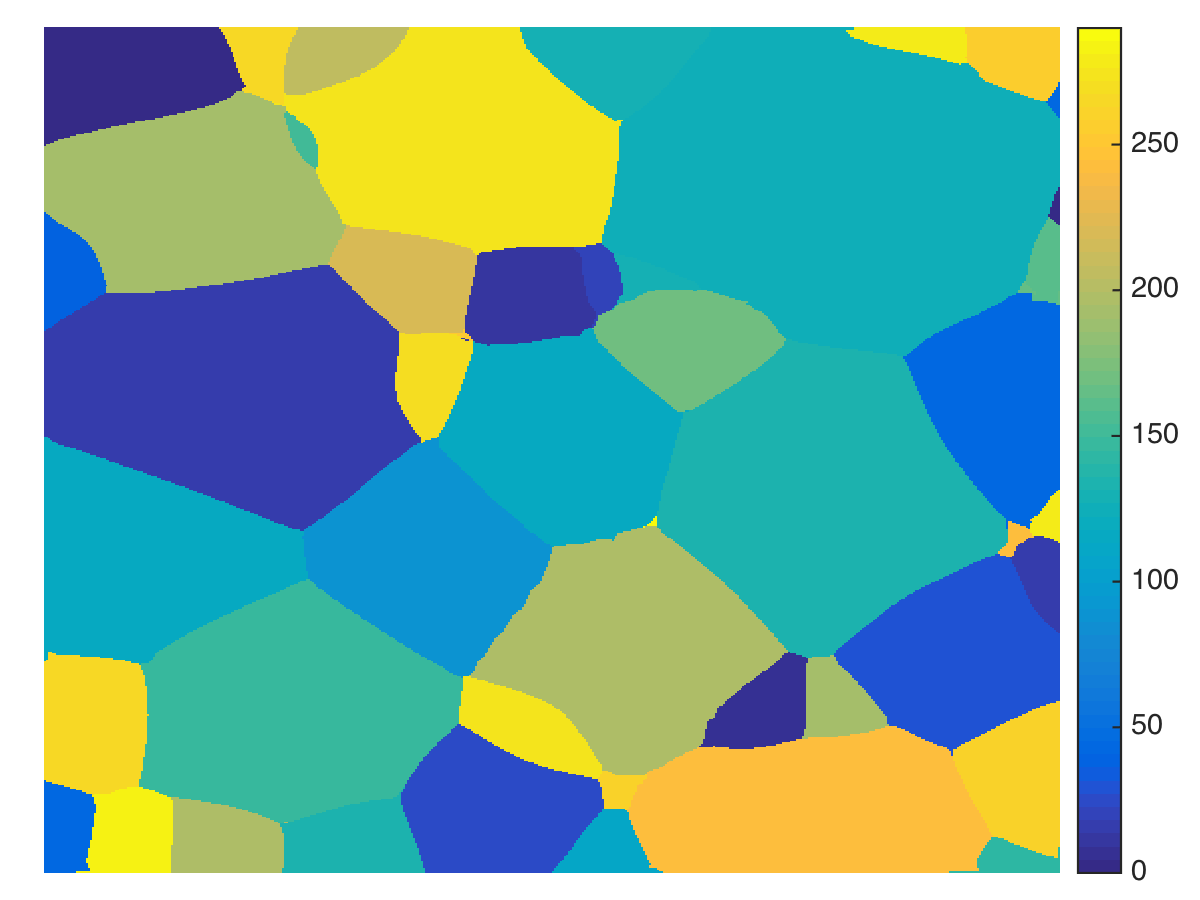}}\\

\hspace*{1.45in} \subfigure[{ \scriptsize RS: Distortion Image  \hspace{5mm} (TD  $ = 2.11 \times 10^{-2} $)}]{\includegraphics[height=1.3in,width=1.4in]{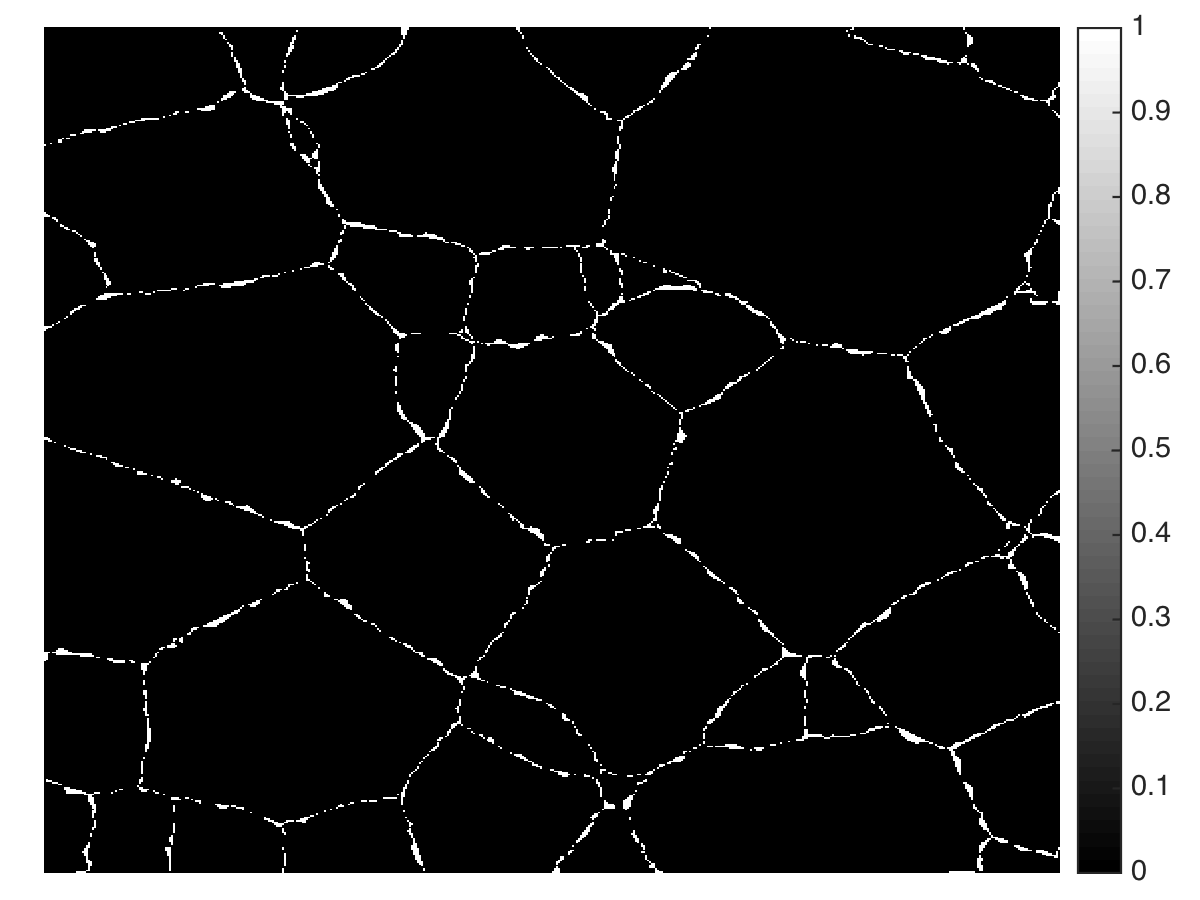}}
 \subfigure[{ \scriptsize LS: Distortion Image \hspace{5mm}  (TD $ = 1.74 \times 10^{-2}$)}]{\includegraphics[height=1.3in,width=1.4in]{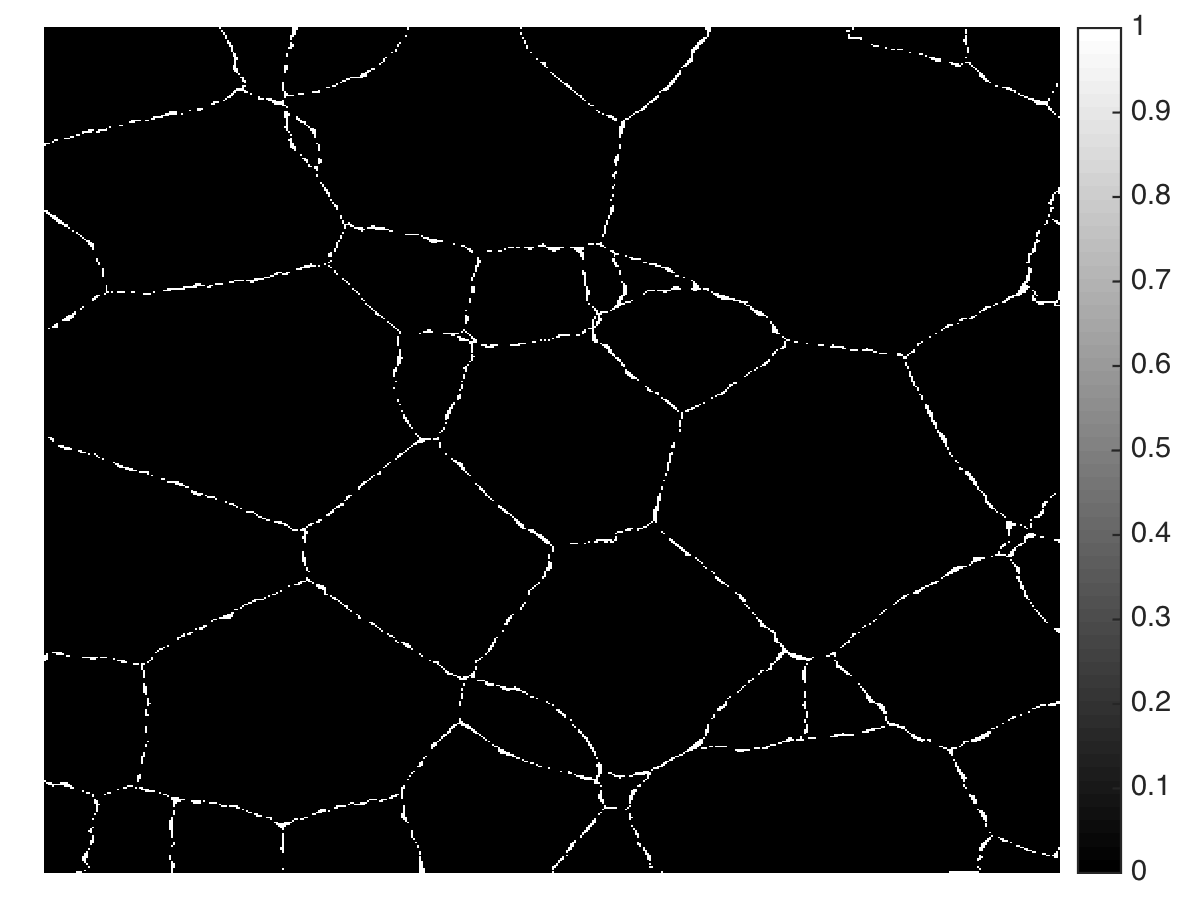}}
 \subfigure[{ \scriptsize SLADS: Distortion Image \hspace{5mm}  (TD $ = 3.81 \times 10^{-6}$)}]{\includegraphics[height=1.3in,width=1.4in
]{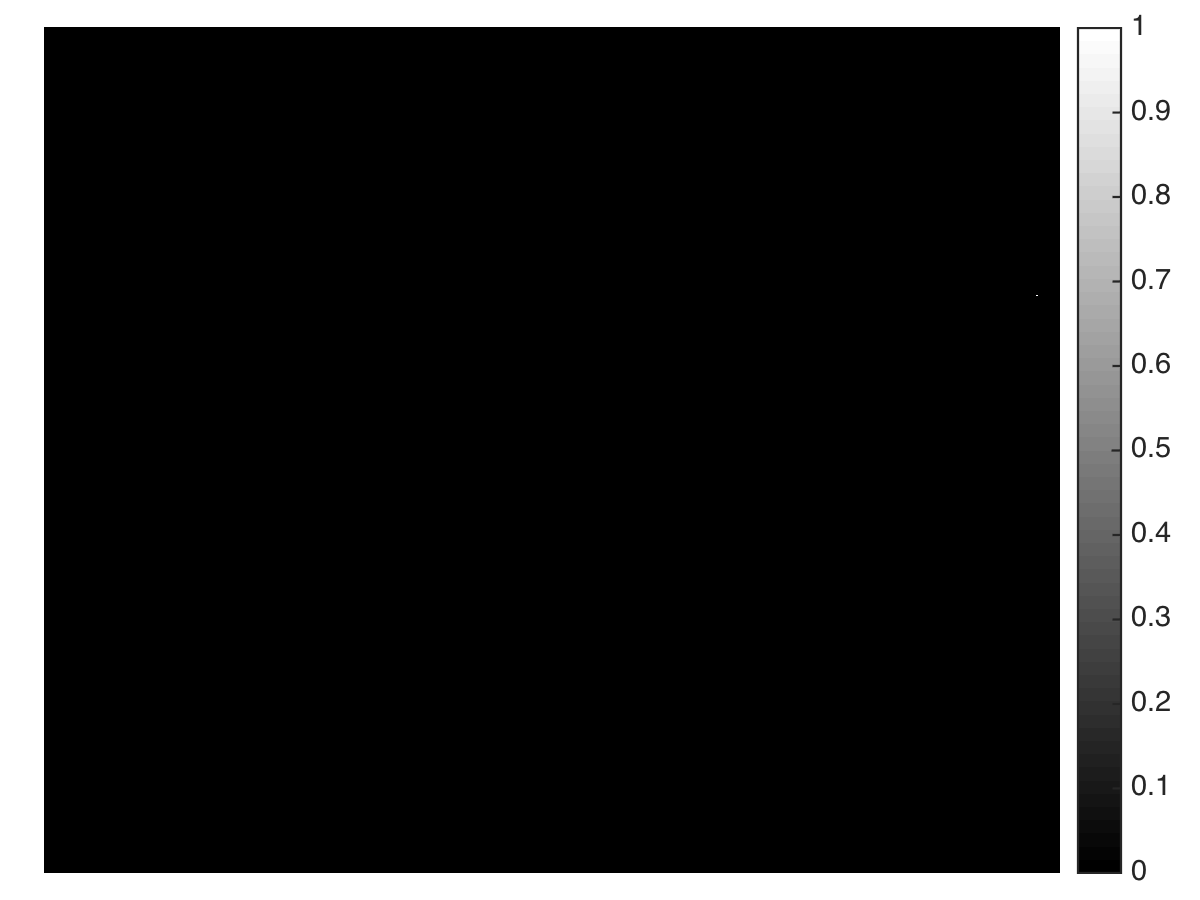}}

\caption{This figure shows sample locations and reconstructions 
after $6\%$ of the image in (a) was sampled using SLADS, RS and LS.
Here (a) is the image being sampled. 
(b), (c) and (d) show the images reconstructed
from samples collected using RS, LS and SLADS respectively. 
(e), (f) and (g) are the distortion images
that correspond to (b), (c) and (d) respectively.
A distortion image is defined as $D \left( X,\hat{X} \right)$ 
where $X$ is the ground truth and the $\hat{X}$ is the reconstructed image.
Note that the distortion image only has values $0$ and $1$
since these are discretely labeled images (see Appendix \ref{sec: distortion metrics}).
(h), (i) and (j) are the measurement masks
that correspond to (b), (c) and (d) respectively.}
\label{fig:discrete results one image}
\end{figure*}

\begin{figure}
\centering
\includegraphics[height=2.3in,width=3in]{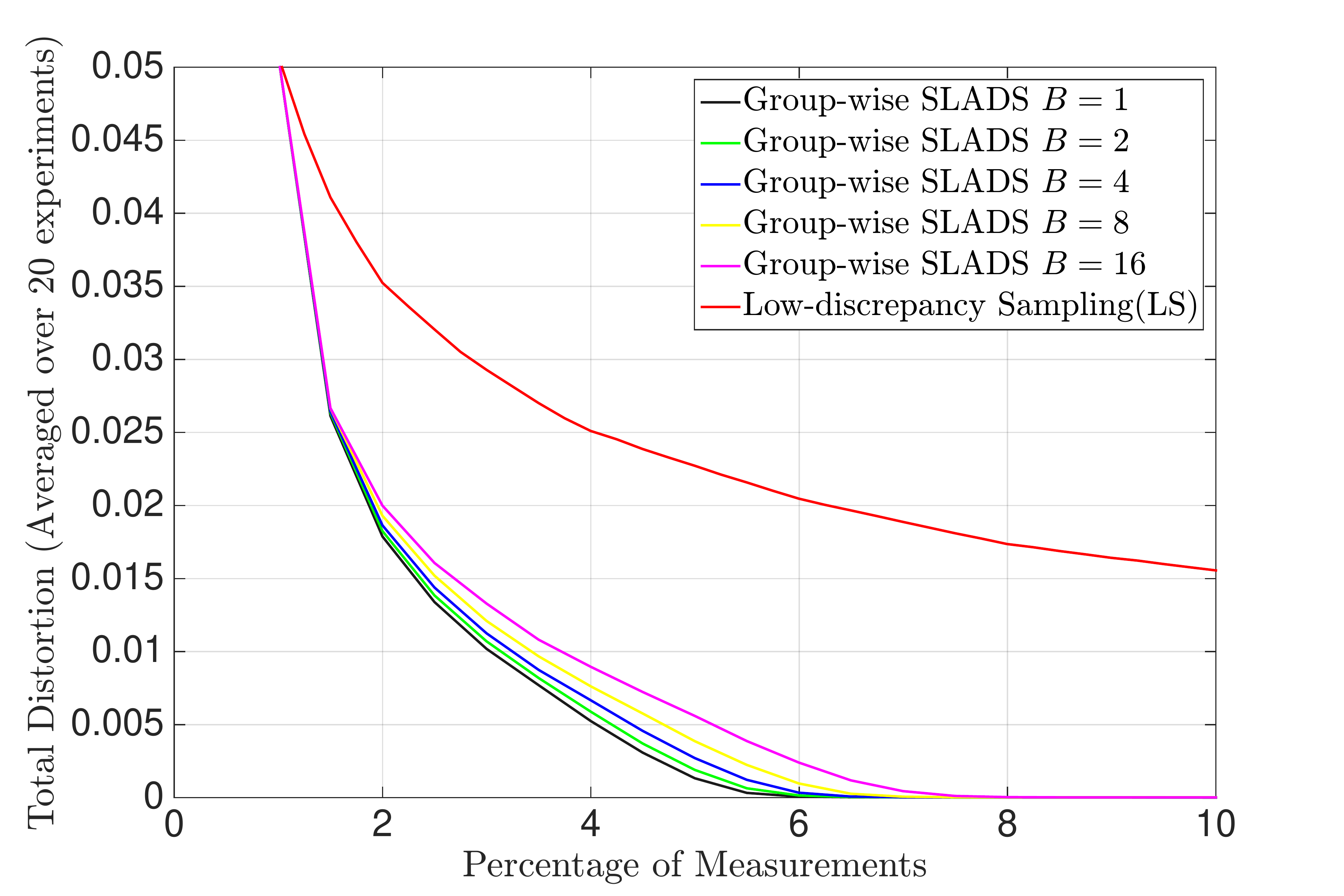}
\caption{In this figure we compare SLADS with Group-wise SLADS. 
In particular, we performed SLADS and Group-wise SLADS with $B=2,4,8,16$ on $20$ simulated EBSD images shown in Figure~\ref{fig:discrete images}(b) and then plotted $TD_k$ versus the percentage of samples.
We see that Group-wise SLADS is comparable to SLADS
and that as $B$ gets larger the distortion for the same number of samples gets larger. 
Note that SLADS is group-wise SLADS with $B=1$.
}
\label{fig:Discrete-GS TD plot}
\end{figure}

\subsection{Results using Simulated EBSD Images}
\label{sec:SimulatedEBSDExperiments}

In this section we first compare SLADS
with two static sampling methods -- Random Sampling (RS) \cite{Hyrum13} and Low-discrepancy Sampling (LS) \cite{LDSampling}.
Then we evaluate the group-wise SLADS method introduced in Section~\ref{sec:group-wise SLADS}
and finally we evaluate the stopping method introduced in Section~\ref{sec:stopping condition}.
Figures~\ref{fig:discrete images}(a) and~\ref{fig:discrete images}(b) show the simulated $512 \times 512$ EBSD images
we used for training and testing, respectively, for all the experiments in this section.
All results used the parameter value $c^*=10$ that was estimated using the method described in Section~\ref{sec:selecting c}.
The average total distortion, $\overline{TD}_k$, for the experiments was computed over the full test set of images.

Figure~\ref{fig:discrete TD plot} shows a plot of the average total distortion, $\overline{TD}_k$,
for each of the three algorithms that were compared, LS, RS and SLADS.
Notice that SLADS dramatically reduces error relative to LS or RS at the same percentage of samples,
and that it achieves nearly perfect reconstruction after approximately $6\%$ of the samples are measured.

Figure~\ref{fig:discrete results one image} gives some insight into the methods
by showing the sampled pixel locations after $6\%$ of samples have been taken for each of the three methods.
Notice that SLADS primarily samples in locations along edges,
but also selects some points in uniform regions.
This both localizes the edges more precisely while also reducing the possibility of missing
a small region or ``island'' in the center of a uniform region.
Alternatively, the LS and RS algorithms select sample locations independent
of the measurements;
so samples are used less efficiently, and the resulting reconstructions have substantially more errors along boundaries.

To evaluate the group-wise SLADS method 
we compare it with SLADS and LS.
Figure~\ref{fig:Discrete-GS TD plot} shows a plot of the average total distortion, $\overline{TD}_k$,
for SLADS, LS, group-wise SLADS with the group sampling rates of $B=2,4,8$ and $16$ performed on the images in 
Figure~\ref{fig:discrete images}(b).
We see that group-wise SLADS has somewhat higher distortion for the same number
of samples as SLADS
and that the distortion increases with increasing values of $B$. 
This is reasonable since SLADS without group sampling has the advantage 
of having the most information available when choosing each new sample.
However, even when collecting $B=16$ samples in a group,
the increase in distortion is still dramatically reduced relative to LS sampling. 

We then evaluate the stopping method
by attempting to stop SLADS at different distortion levels.
In particular, we will attempt to stop SLADS 
when $TD_k \leq TD_{desired}$ for $TD_{desired} = \left\lbrace 5 \times 10^{-5}, 10 \times 10^{-5}, 15 \times 10^{-5} \hdots 50 \times 10^{-5} \right\rbrace$.
For each $TD_{desired}$ value
we found the threshold to place on the stopping function, in equation \ref{eqn:stopping condition},
by using the method described in Section~\ref{sec:stopping condition} 
on a subset of the images in Figure~\ref{fig:discrete images}(a).
Again we used the images shown in Figures~\ref{fig:discrete images}(a) and~\ref{fig:discrete images}(b) 
for training and testing, respectively.
After each SLADS experiment stopped 
we computed the true $TD$ value, ${TD}_{true}$,
and then computed the average true $TD$ value
for a given $TD_{desired}$, 
$\bar{TD}_{true} \left( TD_{desired} \right)$,
by averaging the ${TD}_{true}$ values 
over the $20$ testing images. 

\begin{figure}
\centering
\includegraphics[height=2.3in,width=3in]{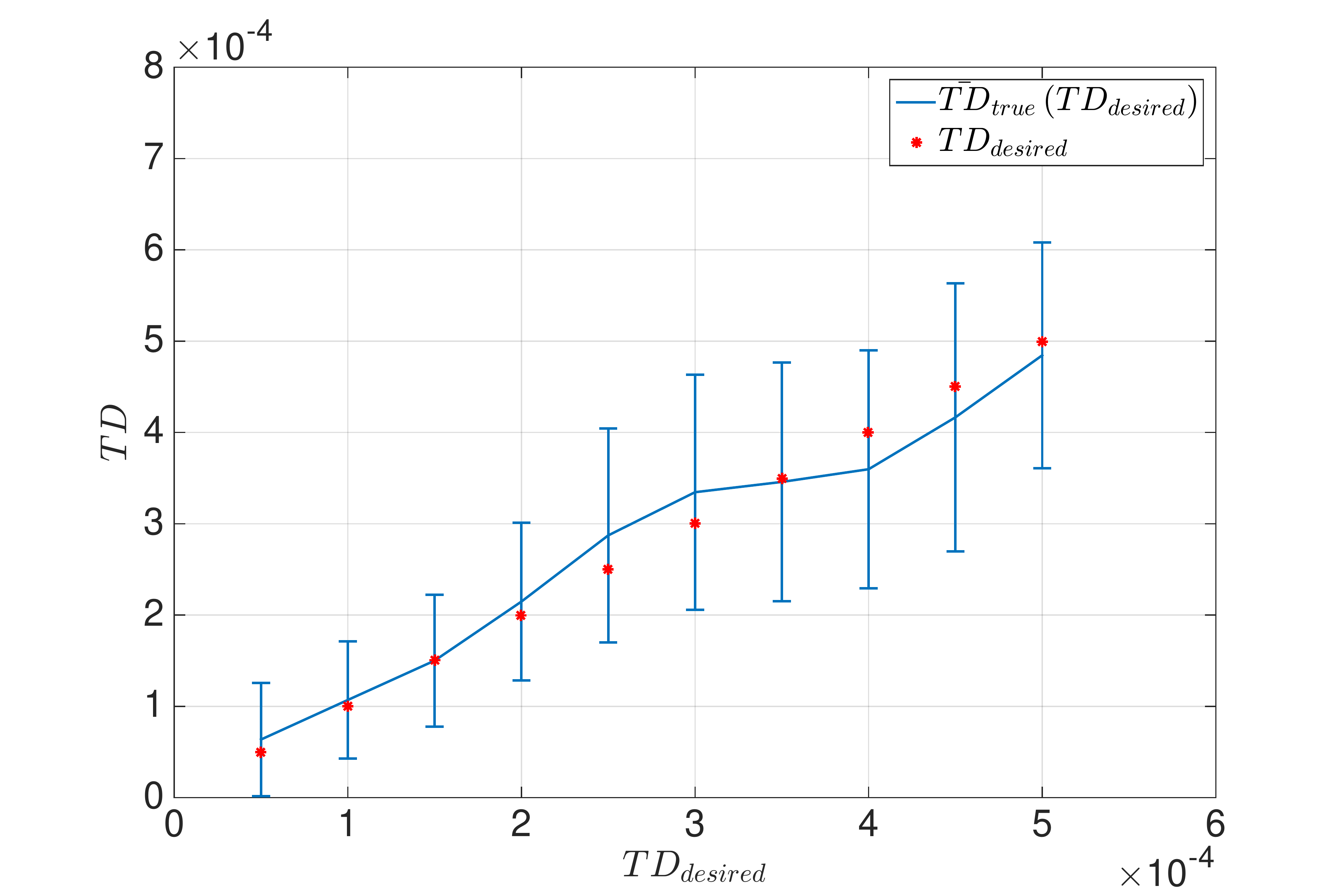}
\caption{This plot corresponds to the experiment performed to evaluate the stopping method we presented in Section~\ref{sec:stopping condition}.
Here we plot $\bar{TD}_{true} \left( TD_{desired} \right)$ 
and $TD_{desired}$ 
where $TD_{desired} = \left\lbrace 5 \times 10^{-5}, 10 \times 10^{-5}, 15 \times 10^{-5} \hdots 50 \times 10^{-5} \right\rbrace$.
$\bar{TD}_{true} \left( TD_{desired} \right)$ was computed
by averaging the ${TD}_{true}$ values 
over $10$ experiments.
Note that we also show the standard deviation bars for $\bar{TD}_{true} \left( TD_{desired} \right)$.
From this plot we see that the stopping method on average is accurate.
}
\label{fig:stop eval}
\end{figure}

Figure~\ref{fig:stop eval} shows a plot of $\bar{TD}_{true} \left( TD_{desired} \right)$ 
and $TD_{desired}$.
From this plot we can see that the experiments 
that in general $TD_{desired} \geq \bar{TD}_{true} \left( TD_{desired} \right)$.
This is the desirable result 
since we intended to stop when $TD_k \leq TD_{desired}$.
However, from the standard deviation bars 
we see that in certain experiments the deviation from $TD_{desired}$ is somewhat high 
and therefore note the need for improvement through future research.

It is also important to mention that the SLADS algorithm (for discrete images) was implemented for protein crystal positioning by Simpson et al. in the synchrotron facility at the Argonne national laboratory \cite{Garth}.

\subsection{Results using Scanning Electron Microscope Images}
\label{sec:MeasuredSEMExperiments}

\begin{figure*}
\centering
\subcapraggedrighttrue
\subfigure[{ \scriptsize Original Image}]{\includegraphics[height=1.3in,width=1.4in]{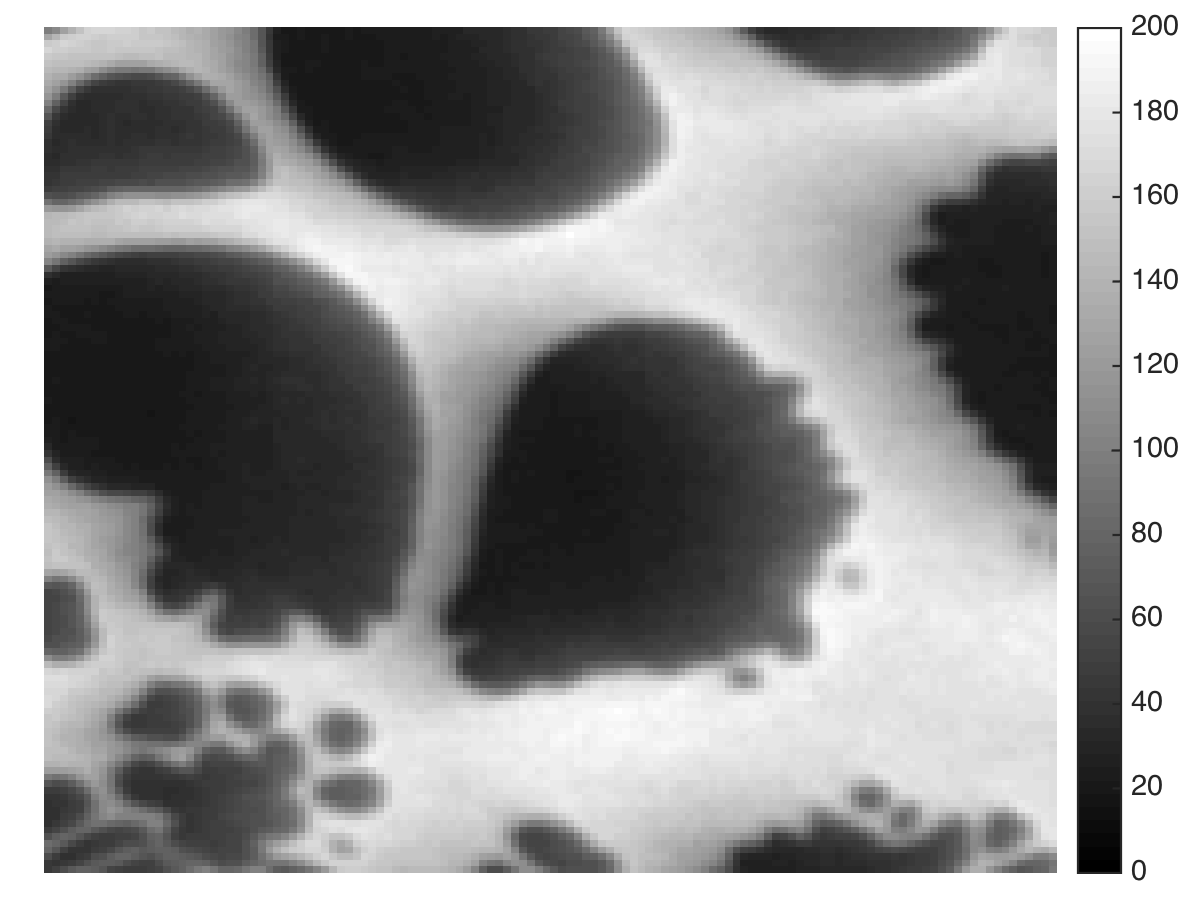}} 
\subfigure[{ \scriptsize RS: Sample locations \hspace{5mm}  ($\sim 15 \%$) }]{\includegraphics[height=1.3in,width=1.4in
]{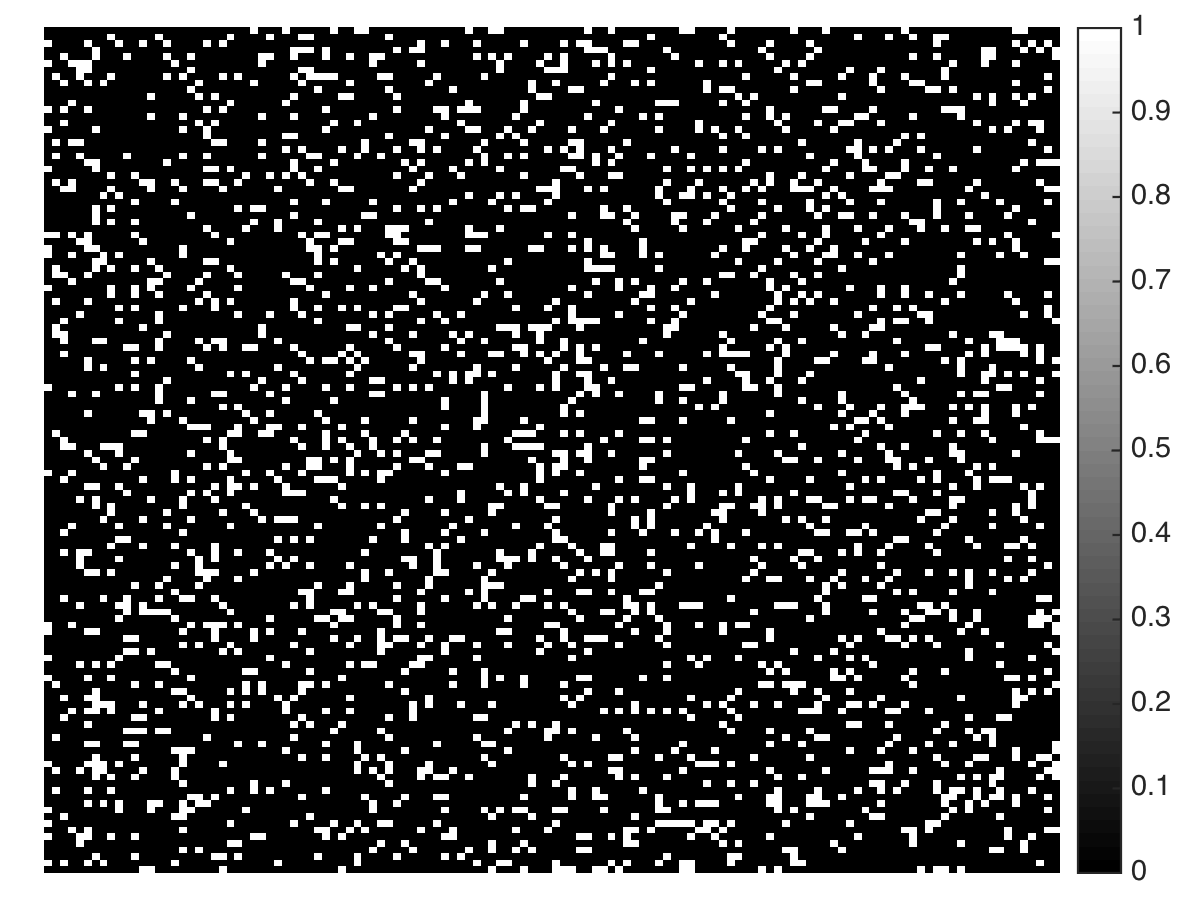}}
 \subfigure[{ \scriptsize LS: Sample locations \hspace{5mm}   ($\sim 15 \%$) }]{\includegraphics[height=1.3in,width=1.4in
]{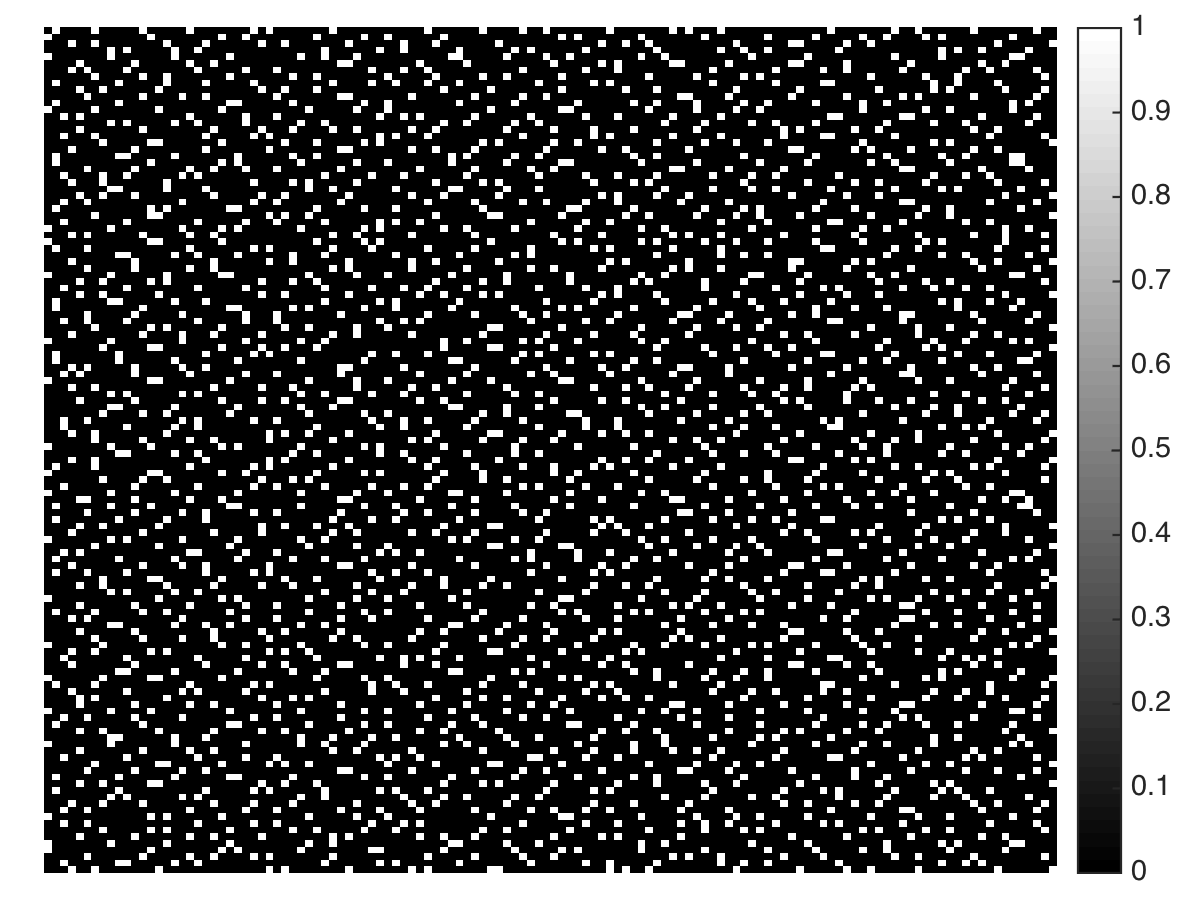}}
 \subfigure[{ \scriptsize SLADS: Sample locations  ($\sim 15 \%$) }]{\includegraphics[height=1.3in,width=1.4in
]{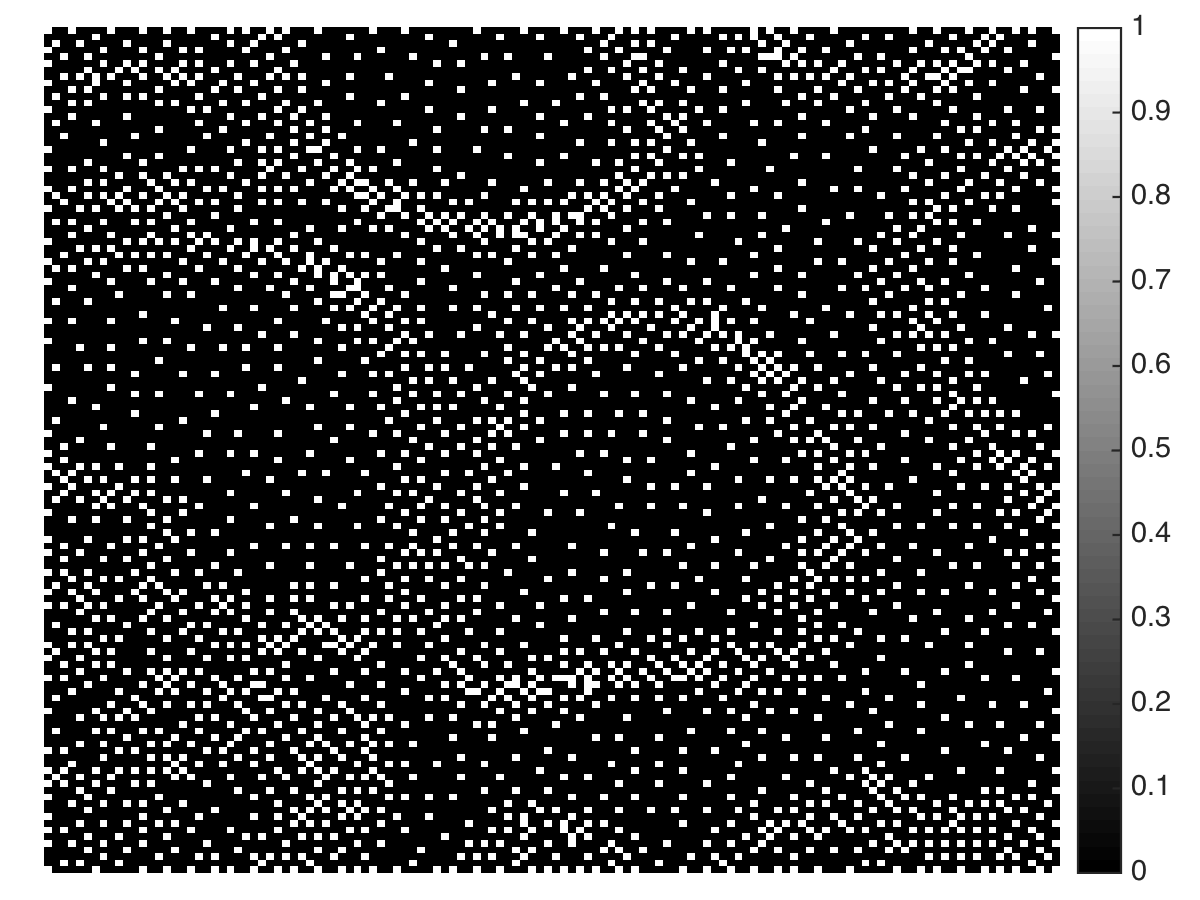}}\\

\hspace*{1.45in} \subfigure[{ \scriptsize RS: Reconstructed Image}]{\includegraphics[height=1.3in,width=1.4in]{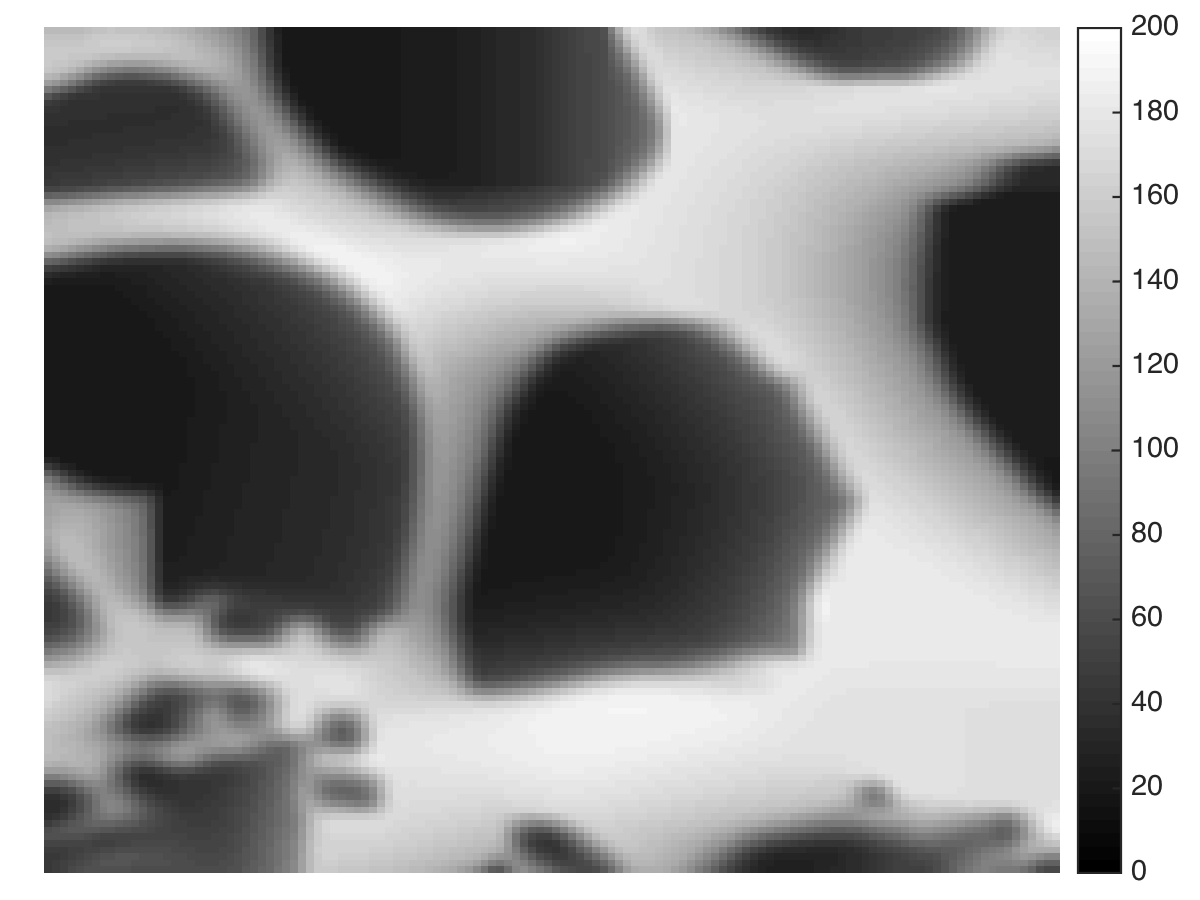}}
 \subfigure[{ \scriptsize LS: Reconstructed Image}]{\includegraphics[height=1.3in,width=1.4in]{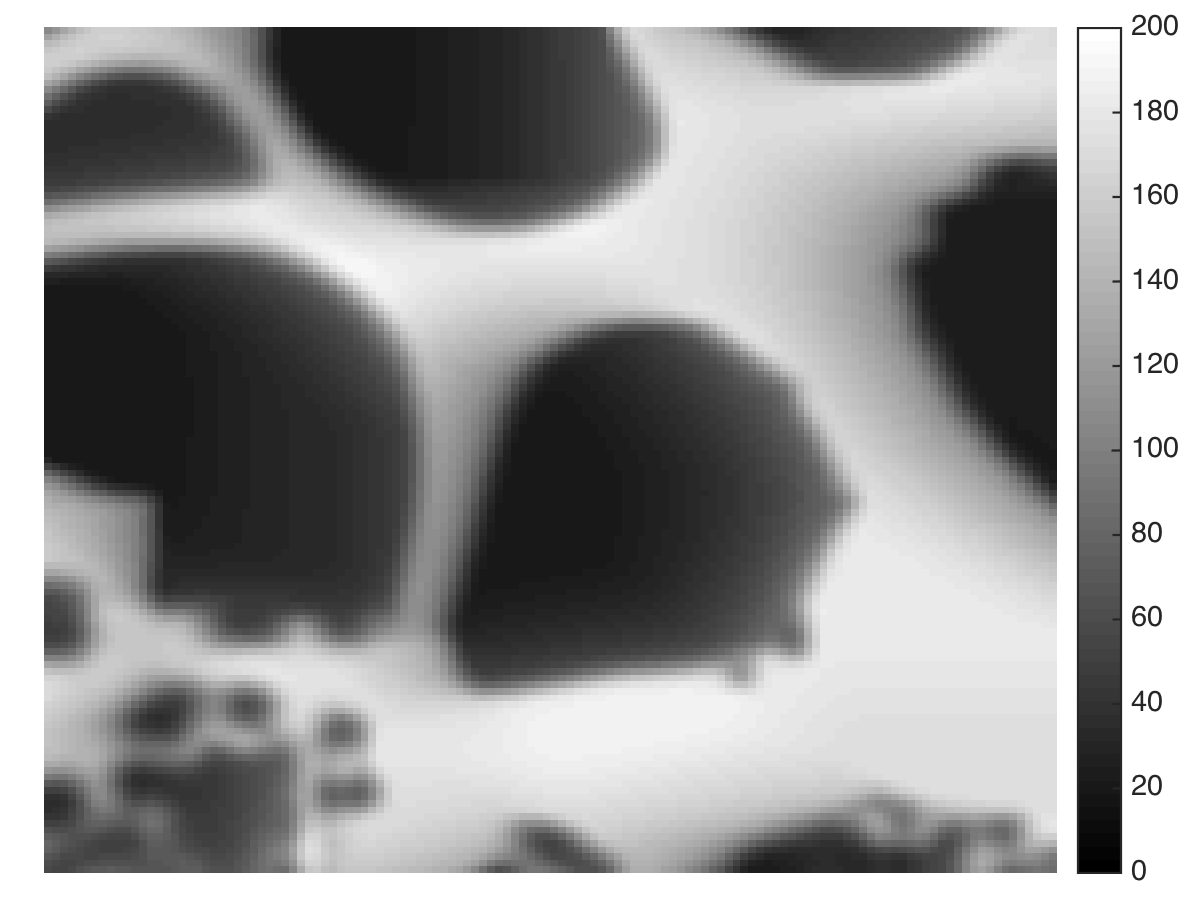}}
 \subfigure[{ \scriptsize SLADS: Reconstructed Image}]{\includegraphics[height=1.3in,width=1.4in
]{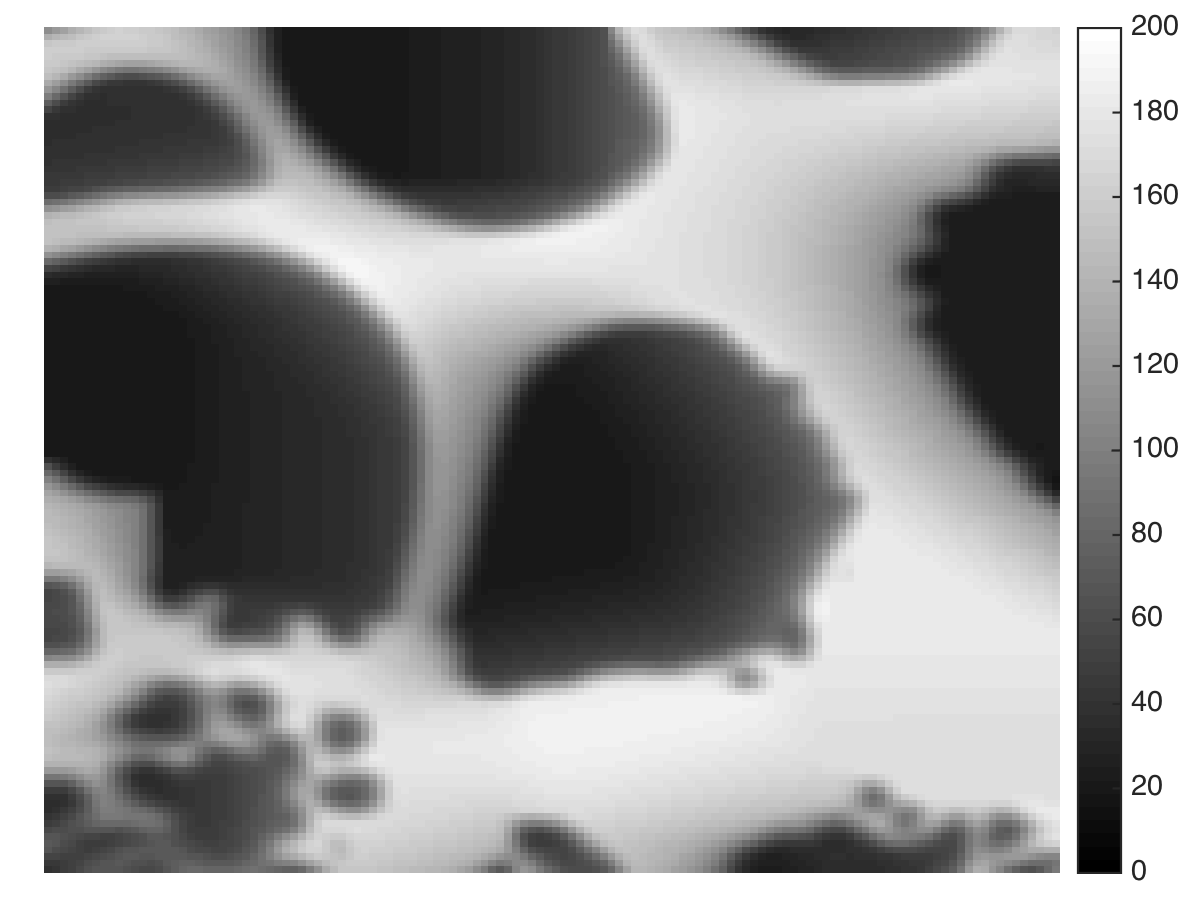}}\\

\hspace*{1.45in} \subfigure[{ \scriptsize RS: Distortion Image  \hspace{5mm} (TD  $=3.88 $)}]{\includegraphics[height=1.3in,width=1.4in]{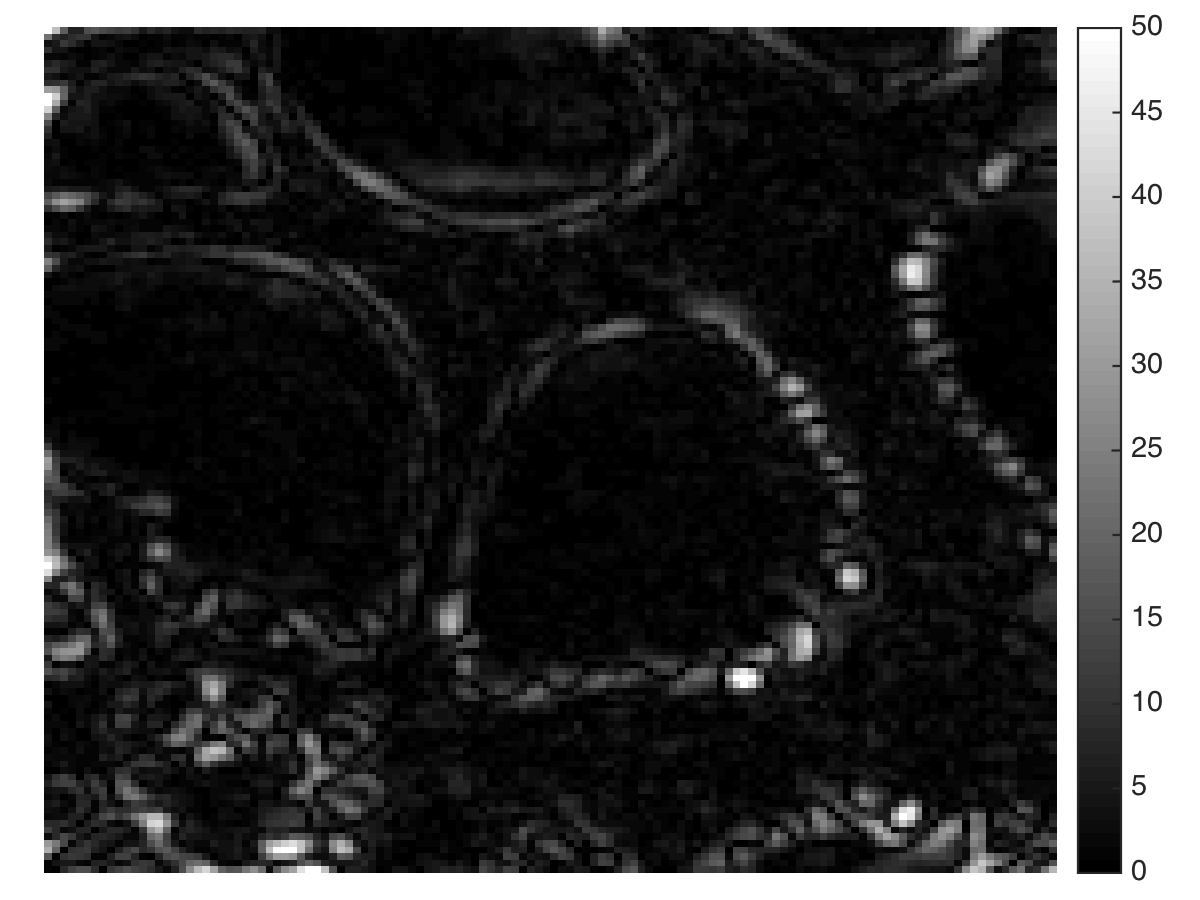}}
 \subfigure[{ \scriptsize LS: Distortion Image \hspace{5mm}  (TD $= 3.44$)}]{\includegraphics[height=1.3in,width=1.4in]{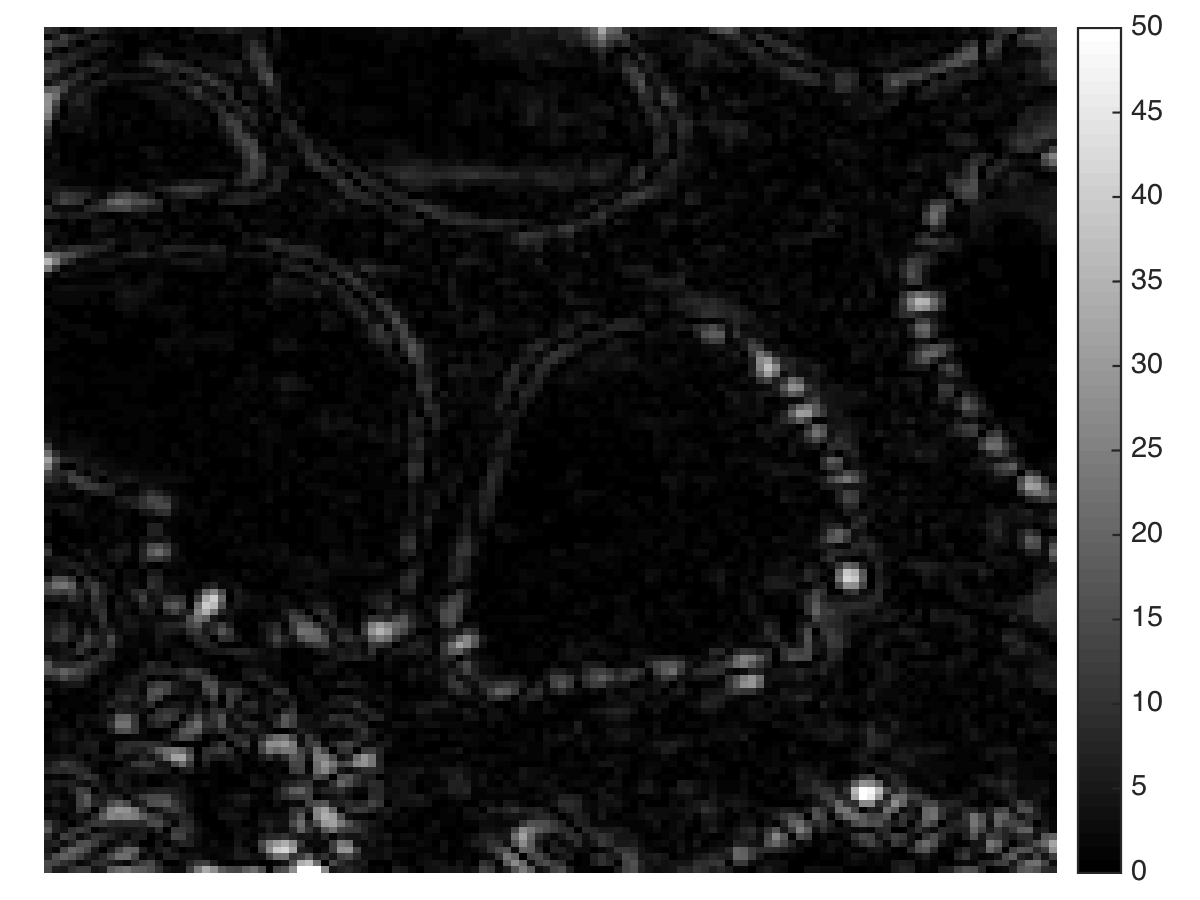}}
 \subfigure[{ \scriptsize SLADS: Distortion Image \hspace{5mm}  (TD $=2.63$)}]{\includegraphics[height=1.3in,width=1.4in
]{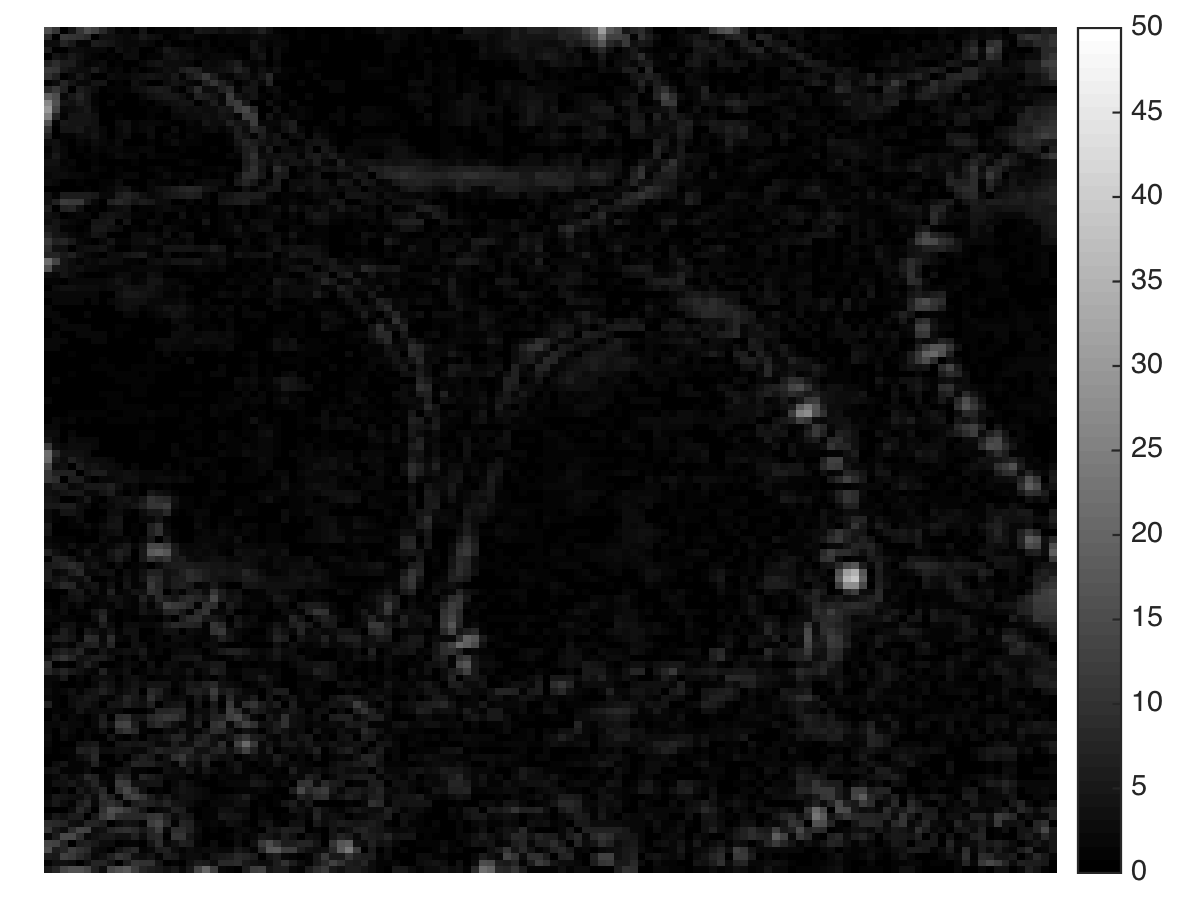}}\\

\caption{This figure shows sample locations and reconstructions 
after $15\%$ of the image in (a) was sampled using SLADS, RS and LS.
Here (a) is the image being sampled. 
(b), (c) and (d) show the images reconstructed
from samples collected using RS, LS and SLADS respectively. 
(e), (f) and (g) are the distortion images
that correspond to (b), (c) and (d) respectively.
A distortion image is defined as $D \left( X,\hat{X} \right)$ 
where $X$ is the ground truth and the $\hat{X}$ is the reconstructed image.
Note that the distortion image has values ranging from $0$ to $255$
since these are continuous $8$ bit images (see Appendix \ref{sec: distortion metrics}).
(h), (i) and (j) are the measurement masks
that correspond to (b), (c) and (d) respectively.}
\label{fig:continuous results one image}
\end{figure*}

\begin{figure}
\centering
\subfigure[]{\includegraphics[height=1.3in,width=1.3in]{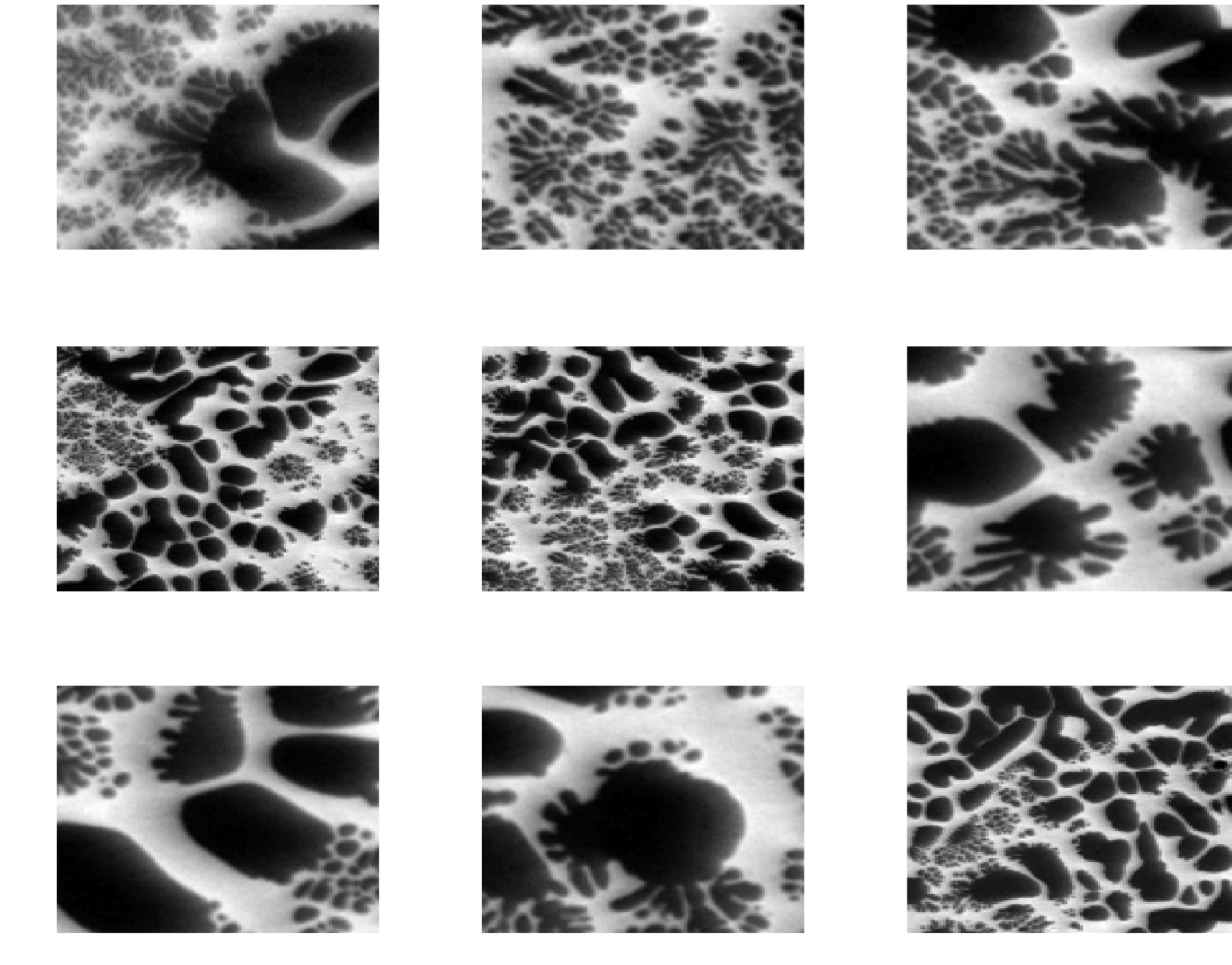}} 
\hspace{20mm}
\subfigure[]{\includegraphics[height=.85in,width=0.85in]{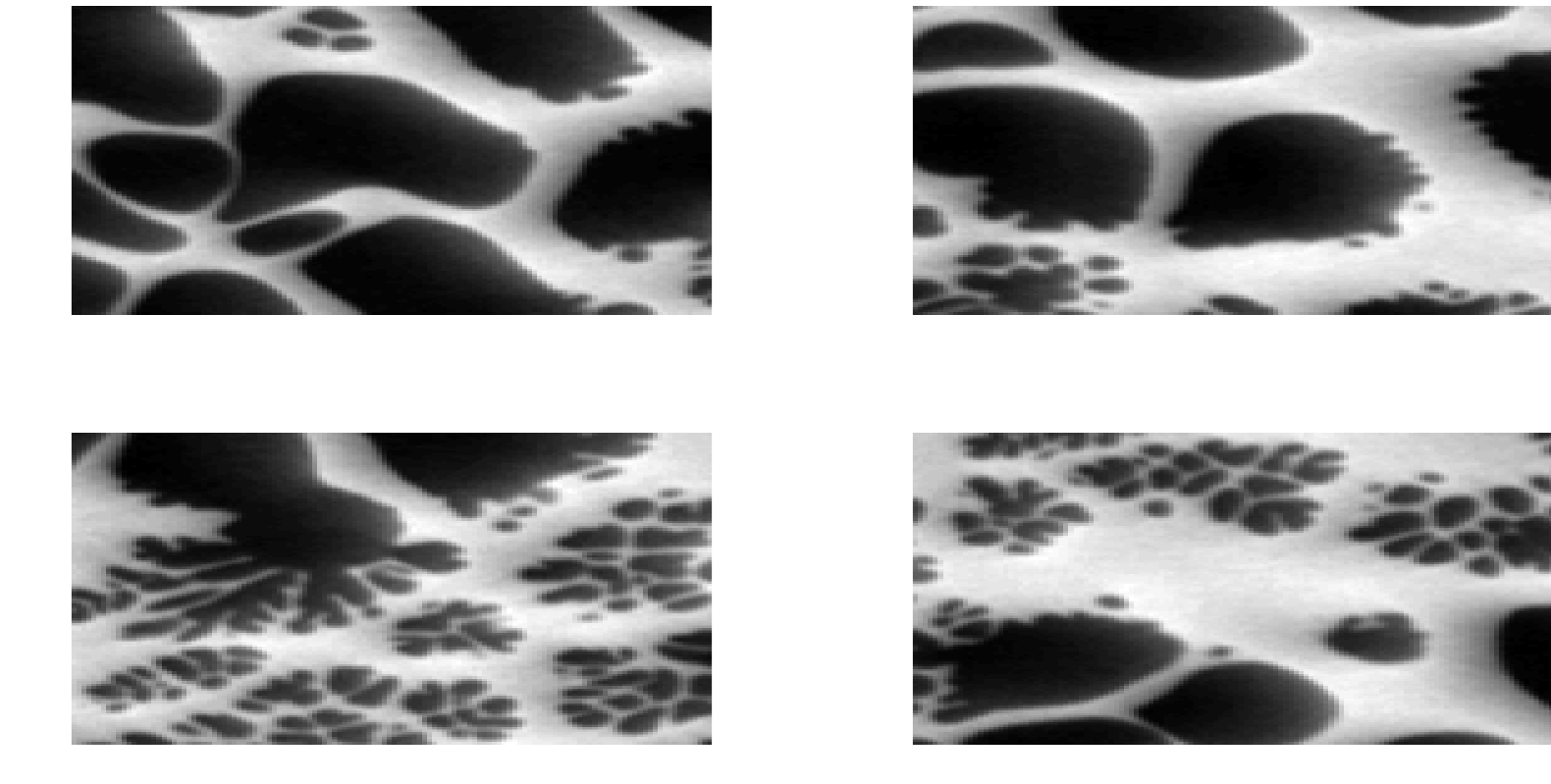}}
\caption{Images used for training and testing in the experiment detailed in Section \ref{sec:MeasuredSEMExperiments} in which we compared SLADS with LS and RS. 
These images have $128 \times 128$ pixels each,
and experimental collected SEM images. 
In particular, (a) shows the images that were used for training and 
(b) shows the images that were used for testing.
These images were collected by Ali Khosravani \& Prof. Surya Kalidindi 
from Georgia Institute of Technology.
}
\label{fig:continuous images}
\end{figure}

\begin{figure}
\centering
\includegraphics[height=2.3in,width=3in]{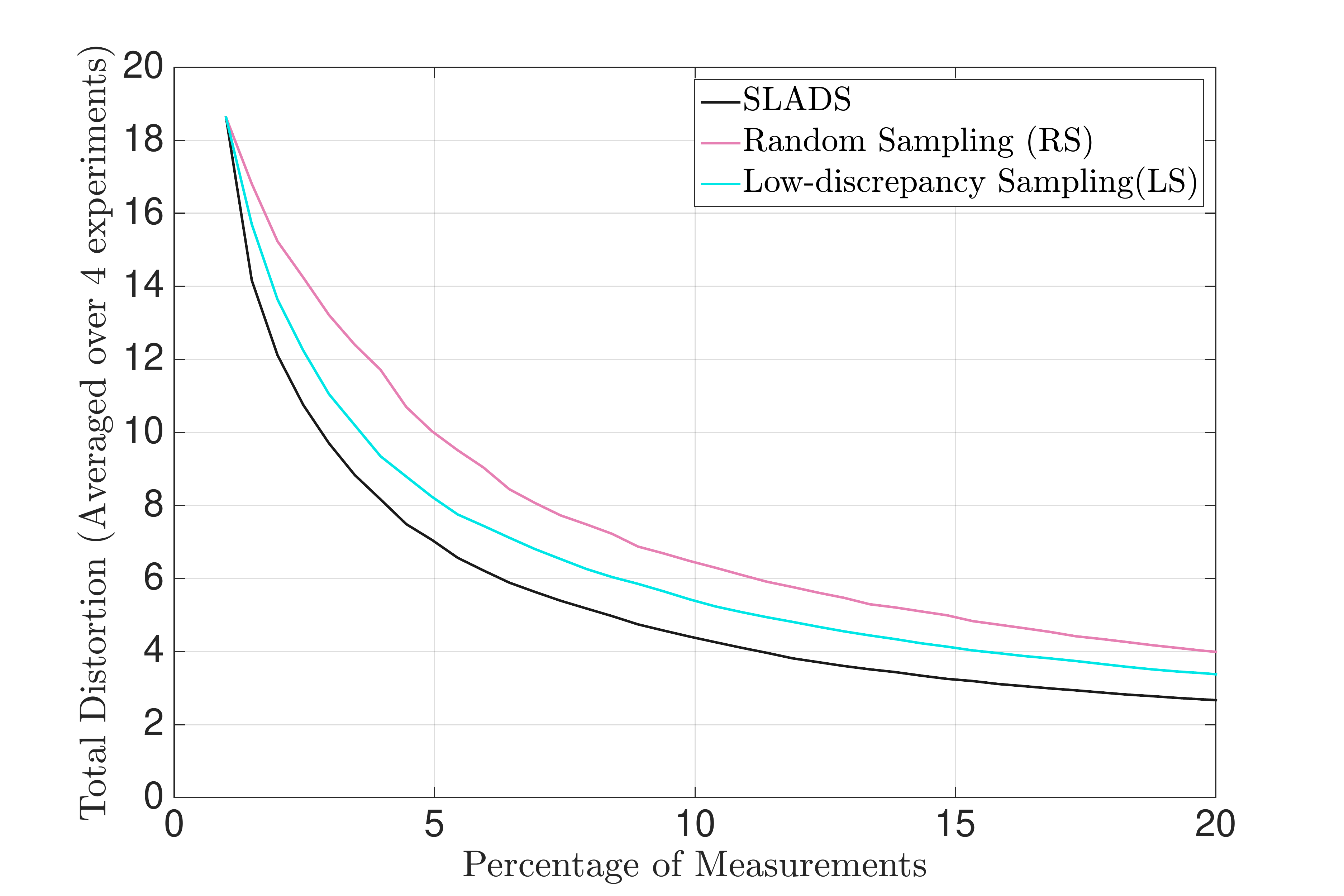}
\caption{In this figure we plot the average $TD$,$\bar{TD}_k$, (averaged over $4$ sampling experiments) 
versus the percentage of samples 
for the experiment detailed in Section~\ref{sec:MeasuredSEMExperiments}.
The plot shows the $\bar{TD}_k$ curves 
that correspond to SLADS, RS and LS.
The images that were sampled in this experiment are shown in Figure~\ref{fig:continuous images}(b).
}
\label{fig:continuous TD plot}
\end{figure}

In this section we again compare SLADS
with LS and RS but now on continuously valued scanning electron microscope (SEM) images. 
Figures~\ref{fig:continuous images}(a) and~\ref{fig:continuous images}(b) show the $128 \times 128$ SEM images
used for training and testing, respectively.
Using the methods described in Section~\ref{sec:selecting c},
the parameter value $c^*=2$ was estimated
and again the average total distortion, $\overline{TD}_k$ was computed over the full test set of images.

Figure~\ref{fig:continuous TD plot} shows a plot of $\overline{TD}_k$ 
for each of the three tested algorithms, SLADS, RS, and LS.
We again see that SLADS outperforms the static sampling methods,
but not as dramatically as in the discrete case.

Figure~\ref{fig:continuous results one image} shows the results
of the three sampling methods after $15\%$ of the samples have been taken
along with the resulting sampling locations that were measured.
Once more we notice that SLADS primarily samples along edges,
and therefore we get better edge resolution.
We also notice that some of the smaller dark regions (``islands'') are missed by LS and RS 
while SLADS is able to resolve almost all of them. 

\section{Conclusions}

In this paper, we presented a framework 
for dynamic image sampling which we call 
a supervised learning approach to dynamic sampling (SLADS).
The method works by selecting the next measurement location 
in a manner that minimizes the expected reduction in distortion (ERD)
for each new measurement.
The SLADS algorithm dramatically reduces the computation required for dynamic sampling 
by using a supervised learning approach in which a regression algorithm
is used to efficiently estimate the ERD for each new measurement.
This makes the SLADS algorithm practical for real-time implementation.

Our experiments show that SLADS can dramatically outperform static sampling methods 
for the measurement of discrete data.
For example, SEM analytical methods such as EBSD\cite{EBSD},
or synchrotron crystal imaging \cite{Garth} are just two cases in which sampling of discrete images is important.
We also introduced a group-wise SLADS method which allows for sampling of multiple pixels in a group,
with only limited loss in performance.
Finally, we concluded with simulations on sampling from continuous SEM images in which we demonstrated that SLADS provides modest improvements compared to static sampling.

\begin{appendix}

\subsection{Distortion Metrics for Experiments}
\label{sec: distortion metrics}
Applications such as EBSD generate images formed by discrete classes.
For these images, we use a distortion metric defined
between two vectors $A \in \mathbb{R}^N$ and $B\in \mathbb{R}^N$ as
\begin{equation}
D\left(A,B \right) = \displaystyle\sum\limits_{i=1}^N I \left( A_i,B_i \right),
\end{equation}
where $I$ is an indicator function defined as 
\begin{equation}
I \left( A_i,B_i \right) =\begin{cases}
0 \quad  A_i=B_i\\
1 \quad A_i \neq B_i,
\end{cases}
\end{equation}
where $A_i$ is the $i^{th}$ element of the vector $i$.

However, for the experiments in Section~\ref{sec:MeasuredSEMExperiments} we used continuously valued images. 
Therefore, we defined the distortion $D\left( A,B\right)$
between two vectors $A$ and $B$ as
\begin{equation}
D\left(A,B \right) =\displaystyle\sum\limits_{i=1}^N  \vert A_i-B_i \vert.
\end{equation}

\subsection{Reconstruction Methods for Experiments}
\label{sec: Reconstruction methods}
In the experiments with discrete images experiments all the reconstructions were performed
using the weighted mode interpolation method.
The weighted mode interpolation of a pixel $s$ is $X_{\hat{r}}$ for
\begin{equation}
\hat{r}=\arg \max_{r\in \partial s} \left\lbrace  \displaystyle\sum\limits_{t \in \partial s} \left[ \left( 1- D \left( X_r,X_t \right) \right) w_r^{(s)} \right] \right\rbrace,
\end{equation}
where 
\begin{equation}
w_{r}^{(s)} =\frac{\frac{1}{\Vert s-r \Vert^2}}{\displaystyle\sum\limits_{u \in \partial s} \frac{1}{\Vert s-u \Vert^2} }
\end{equation}
and $\vert \partial s \vert =10$.

In the training phase of the experiments on continuously valued data, we performed reconstructions using the Plug \& Play algorithm \cite{sreehari2015advanced} to compute the reduction-in-distortion. 
However, to compute the reconstructions 
for descriptors (both in testing and training)
we used weighted mean interpolation
instead of Plug \& Play to minimize the run-time speed of SLADS. 
We define the weighted mean 
for a location $s$ by
\begin{equation}
\hat{X}_s = \displaystyle\sum\limits_{r \in \partial s} w_{r}^{(s)}  X_r.
\end{equation}
\vspace*{-3mm}
\end{appendix}
\section*{ACKNOWLEDGMENTS}
The authors acknowledge support from the Air Force Office of Scientific Research (MURI - Managing the Mosaic of Microstructure, grant \# FA9550-12-1-0458) and from the Air Force Research Laboratory Materials and Manufacturing directorate (Contract \# FA8650-10-D-5201-0038). The authors also thank Ali Khosravani  \& Prof. Surya Kalidindi, Georgia Institute of Technology for providing the images used for dynamic sampling simulation on an experimentally-collected image.
\bibliographystyle{IEEEtran}
{
\nocite{*}
\footnotesize
\bibliography{Ref}
}

\end{document}